\documentclass[10pt,twocolumn,letter]{IEEEtran}

\usepackage{amsmath,graphicx}
\usepackage{amsfonts}
\usepackage{amssymb}
\usepackage{epsfig}
\usepackage{mathrsfs}
\usepackage{graphicx}
\usepackage{algorithm}
\usepackage{caption}
\usepackage{subcaption}
\usepackage{epstopdf}
\usepackage{balance}
\usepackage{color}
\usepackage{cite}

\newsavebox\myboxA
\newsavebox\myboxB
\newlength\mylenA

\newcommand{\oline}[1]{\mkern 1.5mu\overline{\mkern-1.5mu#1\mkern-1.5mu}\mkern 1.5mu}

\def\mA{\mbox{$\mathbf{A}$}}
\def\mB{\mbox{$\mathbf{B}$}}

\def\mC{\mbox{$\mathbf{C}$}}
\def\mD{\mbox{$\mathbf{D}$}}

\def\mG{\mbox{$\mathbf{G}$}}

\def\mH{\mbox{$\mathbf{H}$}}
\def\mI{\mbox{$\mathbf{I}$}}
\def\mJ{\mbox{$\mathbf{J}$}}
\def\mL{\mbox{$\mathbf{L}$}}
\def\mM{\mbox{$\mathbf{M}$}}
\def\mP{\mbox{$\mathbf{P}$}}
\def\mQ{\mbox{$\mathbf{Q}$}}
\def\mR{\mbox{$\mathbf{R}$}}
\def\mT{\mbox{$\mathbf{T}$}}
\def\mU{\mbox{$\mathbf{U}$}}
\def\mV{\mbox{$\mathbf{V}$}}
\def\mW{\mbox{$\mathbf{W}$}}
\def\mX{\mbox{$\mathbf{X}$}}

\def\mZ{\mbox{$\mathbf{Z}$}}

\def\mSigma{\mbox{$\mathbf{\Sigma} \kern .08em$}}
\def\mLambda{\mbox{$\mathbf{\Lambda} \kern .08em$}}

\newcommand{\E}{{\cal E}}
\newcommand{\G}{{\cal G}}

\newcommand{\V}{{\cal V}}

\def\S{\text{\mbox{${\cal S}$}}}
\def\F{\text{\mbox{${\cal F}$}}}

\def\V{\text{\mbox{${\cal V}$}}}
\def\E{\text{\mbox{${\cal E}$}}}
\def\G{\text{\mbox{${\cal G}$}}}

\def\b0{\text{\mbox{\boldmath $0$}}}

\def\bc{\text{\mbox{\boldmath $c$}}}
\def\bd{\text{\mbox{\boldmath $d$}}}
\def\bee{\text{\mbox{\boldmath $e$}}}

\def\bg{\text{\mbox{\boldmath $g$}}}

\def\bq{\text{\mbox{\boldmath $q$}}}
\def\br{\text{\mbox{\boldmath $r$}}}
\def\bs{\text{\mbox{\boldmath $s$}}}
\def\bu{\text{\mbox{\boldmath $u$}}}
\def\bw{\text{\mbox{\boldmath $w$}}}
\def\bx{\text{\mbox{\boldmath $x$}}}
\def\by{\text{\mbox{\boldmath $y$}}}

\def\buno{\text{\mbox{\boldmath $1$}}}

\newenvironment{proof}[1][Proof]{\noindent \textit{#1.} }{\qedsymbol}
\newcommand{\qedsymbol}{\hspace{\fill}\rule{1.5ex}{1.5ex}}

\begin{document}

\title{Distributed Adaptive Learning of Graph Signals\vspace{.1cm}}

\author{Paolo~Di Lorenzo,~\IEEEmembership{Member,~IEEE}, Paolo Banelli,~\IEEEmembership{Member,~IEEE}, \smallskip\\Sergio Barbarossa,~\IEEEmembership{Fellow,~IEEE}, and Stefania Sardellitti,~\IEEEmembership{Member,~IEEE} \vspace{-0.6cm}
%\\\smallskip
%$^1$ Department of Engineering, University of Perugia, Via G. Duranti 93, 06125, Perugia, Italy\\ \smallskip
%$^2$ Department of Information Engineering, Electronics, and Telecommunications, Sapienza University of Rome, Via Eudossiana 18, 00184, Rome, Italy\\ \smallskip
%Email: \texttt{paolo.dilorenzo@unipg.it, paolo.banelli@unipg.it, sergio.barbarossa@uniroma1.it, stefania.sardellitti@uniroma1.it}
\thanks{
Di Lorenzo and Banelli are with the Dept. of Engineering, University of Perugia, Via G. Duranti 93, 06125, Perugia, Italy; Email: \texttt{paolo.dilorenzo@unipg.it, paolo.banelli@unipg.it}. Barbarossa and Sardellitti are with the Department of Information Engineering, Electronics, and Telecommunications, Sapienza University of Rome, Via Eudossiana 18, 00184, Rome, Italy; E-mail: \texttt{sergio.barbarossa@uniroma1.it, stefania.sardellitti@uniroma1.it}. The work of Paolo Di Lorenzo was supported by the ``Fondazione Cassa di Risparmio di Perugia''.}
}

\maketitle

\begin{abstract}
The aim of this paper is to propose distributed strategies for adaptive learning of signals defined over graphs. Assuming the graph signal to be bandlimited, the method enables distributed reconstruction, with guaranteed performance in terms of mean-square error, and tracking from a limited number of sampled observations taken from a subset of vertices. A detailed mean-square analysis is carried out and illustrates the role played by the sampling strategy on the performance of the proposed method. Finally, some useful strategies for distributed selection of the sampling set are provided. Several numerical results validate our theoretical findings, and illustrate the performance of the proposed method for distributed adaptive learning of signals defined over graphs.
\end{abstract}


\section{Introduction}
\label{sec:intro}

Over the last few years, there was a surge of interest in the development of processing tools for the analysis of signals defined over a graph, or graph signals for short, in view of the many potential applications spanning from sensor networks, social media, vehicular networks, big data or biological networks \cite{shuman2013emerging,sandryhaila2013discrete,sandryhaila2014big}. Graph signal processing (GSP) considers signals defined over a discrete domain having a very general structure, represented by a graph, and subsumes classical discrete-time signal processing as a very simple case. Several processing methods for signals defined over a graph were proposed in \cite{sandryhaila2013discrete}, \cite{sandryhaila2014discrete,narang2012perfect,narang2013compact}, and one of the most interesting aspects is that these analysis tools come to depend on the graph topology. A fundamental role in GSP is of course played by spectral analysis, which passes through the definition of the Graph Fourier Transform (GFT). Two main approaches for GFT have been proposed in the literature, based on the projection of the signal onto the eigenvectors of either the graph Laplacian, see, e.g., \cite{shuman2013emerging}, \cite{pesenson2008sampling}, \cite{zhu2012approximating}, or of the adjacency matrix, see, e.g. \cite{sandryhaila2013discrete}, \cite{chen2015discrete}. The first approach is more suited to handle {\it undirected} graphs and builds on the clustering properties of the graph Laplacian eigenvectors and the minimization of the $\ell_2$ norm graph total variation; the second approach applies also to {\it directed} graphs and builds on the interpretation of the adjacency operator as a graph shift operator, which paves the way for all linear shift-invariant filtering methods for graph signals \cite{Puschel1}, \cite{Puschel2}.
%Finally, a very recent contribution proposes a method to build the graph Fourier basis by minimizing the (directed) graph cut size \cite{sardellitti2016graph}.
%% Uncertainty principle%
%Once the GFT was introduced, an uncertainty principle for graph signals was derived in \cite{agaskar2013spectral} and, more recently \cite{pasdeloup2015toward}, \cite{koprowski2015finite}, \cite{tsitsvero2015signals}. The main goal of these works was to assess the relation between the spread of a signal on the vertices of the graph and on its dual domain, as defined by the GFT. In particular, in \cite{tsitsvero2015signals}, the authors give simple closed form expressions for the fundamental tradeoff between the spread of a signal in the graph and in its transformed domain.

% Sampling %
One of the basic and interesting problems in GSP is the development of a \textit{sampling theory} for signals defined over graphs, whose aim is to recover a bandlimited (or approximately bandlimited) graph signal from a subset of its samples. A seminal contribution was given in \cite{pesenson2008sampling}, later extended in \cite{narang2013signal} and, very recently, in \cite{chen2015discrete}, \cite{tsitsvero2015signals}, \cite{wang2014local}, \cite{marquez2015}, \cite{TsitsveroEusipco15}. Several reconstruction methods have been proposed, either iterative as in \cite{wang2014local}, \cite{narang2013localized}, or single shot, as in \cite{chen2015discrete}, \cite{segarra2015reconstruction}, \cite{tsitsvero2015signals}. Frame-based approaches for the reconstruction of graph signals from subsets of samples have also been proposed in \cite{pesenson2008sampling}, \cite{tsitsvero2015signals}, \cite{wang2014local}. Furthermore, as shown in \cite{chen2015discrete}, \cite{tsitsvero2015signals}, dealing with graph signals, the recovery problem may easily become ill-conditioned, depending on the location of the samples. Thus, for any given number of samples, the sampling set plays a fundamental role in the conditioning of the recovery problem. This makes crucial to search for strategies that optimize the selection of the sampling set over the graph.
% Learning from graph signals %
The theory developed in the last years for GSP was then applied to solve specific learning tasks, such as semi-supervised classification on graphs \cite{sandryhaila2013classification}, graph dictionary learning \cite{thanou2013parametric}, smooth graph signal recovery from random samples \cite{zhou2004regularization,belkin2006manifold,chen2015signal,chen2015signalrecovery}, inpainting \cite{chen2014signal}, denoising \cite{chen2014signaldenoising}, and adaptive estimation \cite{di2016least}.

Almost all previous art considers centralized processing methods for graph signals. In many practical systems, data are collected in a distributed network, and sharing local information with a central processor is either unfeasible or not efficient, owing to the large size of the network and volume of data, time-varying network topology, bandwidth/energy constraints, and/or privacy issues. Centralized processing also calls for sufficient resources to transmit the data back and forth between the nodes and the fusion center, which limits the autonomy of the network, and may raise robustness concerns as well, since the central processor represents a bottleneck and an isolate point of failure. In addition, a centralized solution may limit the ability of the nodes to adapt in real-time to time-varying scenarios. Motivated by these observations, in this paper we focus on distributed techniques for graph signal processing. Some distributed methods were recently proposed in the literature, see, e.g. \cite{chen2015distributed,thanou2015distributed,wang2015distributed}. In \cite{chen2015distributed}, a distributed algorithm for graph signal inpainting is proposed; the work in \cite{thanou2015distributed} considers distributed processing of graph signals exploiting graph spectral dictionaries; finally, reference \cite{wang2015distributed} proposes a distributed tracking method for time-varying bandlimited graph signals, assuming perfect observations (i.e., there is no measurement noise) and a fixed sampling strategy.

 \textit{Contributions of the paper:} In this work, we propose distributed strategies for adaptive learning of graph signals. The main contributions are listed in the sequel.
\begin{enumerate}
  \item We formulate the problem of \textit{distributed learning of graph signals} exploiting a \textit{probabilistic sampling} scheme over the graph;
  \item We provide necessary and sufficient \textit{conditions for adaptive reconstruction} of the signal from the graph samples;
  \item We apply diffusion adaptation methods to solve the problem of learning graph signals in a distributed manner. The resulting algorithm is a generalization of diffusion adaptation strategies where nodes sample data from the graph with some given probability.
  \item We provide a detailed mean square analysis that illustrates the role of the probabilistic sampling strategy on the performance of the proposed algorithm.
  \item We design useful strategies for the \textit{distributed selection} of the (expected) sampling set. To the best of our knowledge, this is the first strategy available in the literature for distributed selection of graph signal's samples.
\end{enumerate}
The work merges, for the first time in the literature, the well established field of adaptation and learning over networks, see, e.g., \cite{Cattivelli-Sayed,lopes2008diffusion,takahashi2010diffusion,Chen-Sayed,dilorenzo2013sparse,sayed2014adaptation,chen2014multitask,chen2015diffusion,chen2017multitask}, with the emerging area of signal processing on graphs, see, e.g., \cite{shuman2013emerging,sandryhaila2013discrete,sandryhaila2014big}. The proposed method exploits the graph structure that describes the observed signal and, under a bandlimited assumption, enables adaptive reconstruction and tracking from a limited number of observations taken over a subset of vertices in a totally distributed fashion. Interestingly, the graph topology plays an important role both in the processing and communication aspects of the algorithm. A detailed mean-square analysis illustrates the role of the sampling strategy on the reconstruction capability, stability, and performance of the proposed algorithm. Thus, based on these results, we also propose a distributed method to select the set of sampling nodes in an efficient manner. An interesting feature of our proposed strategy is that this subset is allowed to vary over time, provided that the \textit{expected} sampling set satisfies specific conditions enabling signal reconstruction. We expect that the proposed tools will represent a key technology for the distributed proactive sensing of cyber physical systems, where an effective control mechanism requires the availability of data-driven sampling strategies able to monitor the overall system by only checking a limited number of nodes.

The paper is organized as follows. In Sec. II, we introduce some basic GSP tools. Sec. III introduces the proposed distributed algorithm for adaptive learning of graph signals, illustrating also the conditions enabling signal reconstruction from a subset of samples. In Sec. IV we carry out a detailed mean-square analysis, whereas Sec. V is devoted to the development of useful strategies enabling the selection of the sampling set in a totally distributed fashion. Then, in Sec. VI we report several numerical simulations, aimed at assessing the validity of the theoretical analysis and the performance of the proposed algorithm. Finally, Sec. VII draws some conclusions.

\section{Graph Signal Processing Tools}

In this section, we introduce some useful concepts from GSP that will be exploited along the paper. Let us consider a graph $\G = (\V, \E)$ composed of $N$ nodes $\V = \{1,2,..., N\}$, along with a set of weighted edges $\E=\{a_{ij}\}_{i, j \in \V}$, such that $a_{ij}>0$, if there is a link from node $j$ to node $i$, or $a_{ij}=0$, otherwise. The adjacency matrix $\mA=\{a_{ij}\}_{i,j=1}^N\in\mathbb{R}^{N\times N}$ is the collection of all the weights $a_{ij}, i, j = 1, \ldots, N$.  The degree of node $i$ is $k_i:=\sum_{j=1}^{N}a_{ij}$, and the degree matrix $\mathbf{K}$ is a diagonal matrix having the node degrees on its diagonal. The Laplacian matrix is defined as:
%\begin{equation}
$\mathbf{L} = \mathbf{K}-\mathbf{A}$.
%\end{equation}
If the graph is {\it undirected}, the Laplacian matrix is symmetric and positive semi-definite, and admits the eigendecomposition $\mathbf{L}=\mathbf{U}\boldsymbol{\Lambda}\mathbf{U}^H$, where $\mathbf{U}$ collects all the eigenvectors of $\mathbf{L}$ in its columns, whereas $\boldsymbol{\Lambda}$ contains the eigenvalues of $\mathbf{L}$. It is well known from spectral graph theory  \cite{Chung1997} that the eigenvectors of $\mL$ are well suited for representing clusters, since they are signal vectors that minimize the $\ell_2$-norm graph total variation.

A signal $\bx$ over a graph $\G$ is defined as a mapping from the vertex set to the set of complex numbers, i.e. $\bx: \V \rightarrow \mathbb{C}$. In many applications, the signal $\bx$ admits a compact representation, i.e., it can be expressed as:
\begin{equation}
\label{x=Us}
\bx=\mU \bs
\end{equation}
where $\bs$ is exactly (or approximately) sparse. As an example, in all cases where the graph signal exhibits clustering features, i.e. it is a smooth function within each cluster, but it is allowed to vary arbitrarily from one cluster to the other, the representation in (\ref{x=Us}) is compact, i.e., $\bs$ is sparse. A key example is cluster analysis in semi-supervised learning, where a constant signal (label) is associated to each cluster \cite{gadde2014active}. The GFT $\bs$ of a signal $\bx$ is defined as the projection onto the orthogonal set of eigenvectors of the Laplacian \cite{shuman2013emerging}, i.e.,
\begin{equation}
\label{GFT}
\mathbf{\bs} = \mathbf{U}^H \bx.
\end{equation}
The GFT has been defined in alternative ways, see, e.g., \cite{shuman2013emerging}, \cite{sandryhaila2013discrete}, \cite{zhu2012approximating}, \cite{chen2015discrete}. In this paper, we basically follow the approach based on the Laplacian matrix, assuming an undirected graph structure, but the theory could be extended to handle directed graphs considering, e.g., a graph Fourier basis as proposed in \cite{sandryhaila2013discrete}. Also, we denote the support of $\bs$ in (\ref{x=Us}) as $\mathcal{F}=\{i\in\{1,\ldots,N\}:s_i\neq0\}$, and the \textit{bandwidth} of the graph signal $\bx$ is defined as the cardinality of $\mathcal{F}$, i.e. $|\mathcal{F}|$.
%The space of all signals whose GFT is exactly supported on the set $\F$ is known as the {\it Paley-Wiener space} for the set $\F$ \cite{pesenson2008sampling}.
Clearly, combining (\ref{x=Us}) with (\ref{GFT}), if the signal $\bx$ exhibits a clustering behavior, in the sense specified above, the GFT is the way to recover the sparse vector $\bs$. Finally, given a subset of vertices $\S \subseteq \V$, we define a vertex-limiting operator as the matrix
\begin{equation}
\label{D}
\mathbf{D}_{\S} = {\rm diag}\{\buno_{\S}\},
\end{equation}
where $\buno_{\S}$ is the set indicator vector, whose $i$-th entry is equal to one, if  $i \in \S$, or zero otherwise.
%In the rest of the paper, whenever there will be no ambiguities in the specification of the set, we will drop the subscript referring to the set. Finally, given a set $\S$, we denote its complement set as $\S_c$, such that $\V=\S \cup \S_c$ and $\S \cap \S_c=\emptyset$. Thus, we define the vertex-limiting projector onto $\S_c$ as $\mathbf{D}_c$.

\section{Distributed Learning of Graph Signals}

We consider the problem of learning a (possibly time-varying) graph signal from observations taken from a subset of vertices of the graph. The problem fits well, e.g., to a wireless sensor network (WSN) scenario, where the nodes are observing a spatial field related to some physical parameter of interest. Let us assume that the field is either fixed or slowly varying over both the time domain and the vertex (space) domain. Suppose now that the WSN is equipped with nodes that, at every time instant, can decide wether to take (noisy) observations of the underlying signal or not, depending on, e.g., energy constraints, failures, limited memory and/or processing capabilities, etc. Our purpose is to build adaptive techniques that allow the recovery of the field values at each node, pursued using recursive and distributed techniques as new data arrive. In this way, the information is processed on the fly by all nodes and the data diffuse across the network by means of a real-time sharing mechanism.

Let us consider a signal $\bx^o=\{x^o_i\}_{i=1}^N\in\mathbb{C}^N$ defined over the graph $\G = (\V, \E)$. To enable sampling of $\bx^o$ without loss of information, the following is assumed:

\textit{Assumption 1 (Bandlimited): The signal $\bx^o$ is $\F$-bandlimited on the (time-invariant) graph $\mathcal{G}$, i.e., its spectral content is different from zero only on the set of indices $\F$.} \qedsymbol

Under Assumption 1, if the support $\F$ is known beforehand, from (\ref{x=Us}), the graph signal $\bx^o$ can be cast in compact form as:
\begin{equation}
\label{compact_decomp}
\bx^o=\mU_{\F}\bs^o,
\end{equation}
where $\mU_{\F}\in \mathbb{C}^{N\times |\F|}$ collects the subset of columns of matrix $\mU$ in (\ref{x=Us}) associated to the frequency indices $\F$, and $\bs^o\in\mathbb{C}^{|\F|\times 1}$ is the vector of GFT coefficients of the frequency support of the graph signal $\bx^o$. Let us assume that streaming and noisy observations of the graph signal are sampled over a (possibly time-varying) subset of vertices. In such a case, the observation taken by node $i$ at time $n$ can be expressed as:
\begin{align}
\label{lin_observation}
y_i[n]\,=\, d_i[n]\left(x^o_i+v_i[n]\right)= d_i[n]\left( \bc_i^H \bs^o+v_i[n]\right),
\end{align}
$i=1,\ldots,N$, where $^H$ denotes complex conjugate-transposition; $d_i[n]=\{0,1\}$ is a random sampling binary coefficient, which is equal to 1 if node $i$ is taking the observation at time $n$, and 0 otherwise; $v_i[n]$ is a zero-mean, spatially and temporally independent observation noise, with variance $\sigma_i^2$; also, in (\ref{lin_observation}) we have used (\ref{compact_decomp}), where $\bc_i^H\in \mathbb{C}^{1\times|\F|}$ denotes the $i$-th row of matrix $\mU_{\F}$.
In the sequel, we assume that each node $i$ has local knowledge of its corresponding regression vector $\bc_i$ in (\ref{lin_observation}). This is a reasonable assumption even in the distributed scenario considered in this paper. For example, if neighbors in the processing graph can communicate with each other, either directly or via multi-hop routes, there exist many techniques that enable the distributed computation of eigenparameters of matrices describing sparse topologies such as the Laplacian or the adjacency, see, e.g., \cite{kempe2004decentralized,bertrand2013seeing,di2014distributed}. The methods are mainly based on the iterative application of distributed power iteration and consensus methods in order to iteratively compute the desired eigenparameters of the Laplacian (or adjacency) matrix, see, e.g., \cite{di2014distributed} for details. Since we consider graph signals with time-invariant topology, such procedures can be implemented offline during an initialization phase of the network to compute the required regression vectors in a totally distributed fashion. In the case of time-varying graphs, the distributed procedure should be adapted over time, but might result unpractical for large dynamic networks.
%But this case is more challenging and needs further developments for efficient implementations.

The distributed learning task consists in recovering the graph signal $\bx^o$ from the noisy, streaming, and partial observations $y_i[n]$ in (\ref{lin_observation}) by means of in-network adaptive processing. Following a least mean squares (LMS) estimation approach \cite{Cattivelli-Sayed,Chen-Sayed,dilorenzo2013sparse,dilorenzo2014diffusion,sayed2014adaptation}, the task can be cast as the cooperative solution of the following optimization problem:
\begin{align}
\label{diffusion_LMS_problem}
&\min_{\boldsymbol{s}} \;\; \sum_{i=1}^N \;\mathbb{E}_{\,\mathrm{d},\mathrm{v}} \left|d_i[n]\left(y_i[n]-\bc_i^H\bs\right)\right|^2,
\end{align}
where $\mathbb{E}_{\,\mathrm{d},\mathrm{v}}(\cdot)$ denotes the expectation operator evaluated over the random variables $\{d_i[n]\}_{i=1}^N$ and $\{v_i[n]\}_{i=1}^N$, and we have exploited $d_i[n]^2=d_i[n]$ for all $i,n$. In the rest of the paper, to avoid overcrowded symbols, we will drop the subscripts in the expectation symbol referring to the random variables. In the sequel, we first analyze the conditions that enable signal recovery from a subset of samples. Then, we introduce adaptive strategies specifically tailored for the distributed reconstruction of graph signals from a limited number of samples.

\subsection{Conditions for Adaptive Reconstruction of Graph Signals}

In this section, we give a necessary and sufficient condition guaranteeing signal reconstruction from its samples. In particular, assuming the random sampling and observations processes $\bd[n]=\{d_i[n]\}_{i=1}^N$ and $\by[n]=\{y_i[n]\}_{i=1}^N$ to be stationary, the solution of problem (\ref{diffusion_LMS_problem}) is given by the vector $\bs^o$ that satisfies the normal equations:
\begin{align}
\label{normal_equations}
\left(\sum_{i=1}^N \mathbb{E}\{d_i[n]\} \bc_i\bc_i^H \right) \bs^o = \sum_{i=1}^N \mathbb{E}\{d_i[n] y_i[n]\}\,\bc_i .
\end{align}
Letting $p_i=\mathbb{E}\{d_i[n]\}$, $i=1,\ldots,N$, be the probability that node $i$ takes an observation at time $n$, from (\ref{normal_equations}), it is clear that reconstruction of $\bs^o$ is possible only if the matrix
\begin{align}\label{cond}
\sum_{i=1}^N p_i \bc_i\bc_i^H=\mU_{\F}^H\mP\mU_{\F}
\end{align}
is invertible, with $\mP={\rm diag}(p_1,\ldots,p_N)$ denoting a vertex sampling operator as (\ref{D}), but weighted by the sampling probabilities $\{p_i\}_{i=1}^N$.
% the instantaneous sampling set by $\S[n]$, i.e. $\S[n]=\{i=1,\ldots,N \,|\, d_i[n]=1\}$, and
Let us denote the \textit{expected sampling set} by
$$\overline{\S}=\{i=1,\ldots,N \,|\, p_i>0\}.$$
$\overline{\S}$ represents the set of nodes of the graph that collect data with a probability different from zero. From (\ref{normal_equations}) and (\ref{cond}), a necessary condition enabling reconstruction is
\begin{equation}
|\overline{\S}| \geq |\F|,
\end{equation}
i.e., the number of nodes in the expected sampling set must be greater than equal to the signal bandwidth. However, this condition is not sufficient, because matrix $\mU_{\F}^H\mP\mU_{\F}$ in (\ref{cond}) may loose rank, or easily become ill-conditioned, depending on the graph topology and sampling strategy (defined by $\overline{\S}$ and $\mP$).
%Thus, when sampling a graph signal, what matters is not only the number of samples, but also where the samples are taken over the graph.
To provide a condition for signal reconstruction, we proceed similarly to \cite{TsitsveroEusipco15,tsitsvero2015signals,di2016least}. Since $p_i>0$ for all $i\in \overline{\mathcal{S}}$,
\begin{align}\label{cond2}
\mathrm{rank}\left(\sum_{i=1}^N p_i \bc_i\bc_i^H\right)=\mathrm{rank}\left(\sum_{i\in \overline{\mathcal{S}}} \bc_i\bc_i^H\right),
\end{align}
i.e., matrix (\ref{cond}) is invertible if matrix
%\begin{align}
$\sum_{i\in \overline{\mathcal{S}}} \bc_i\bc_i^H=\mU_{\F}^H\mD_{\,\overline{\mathcal{S}}}\mU_{\F}$
%\end{align}
has full rank, where $\mD_{\,\overline{\mathcal{S}}}$ is the vertex-limiting operator that projects onto the expected sampling set $\overline{\S}$. Let us now introduce the operator
\begin{align}\label{exp_compl_oper}
\mD_{\,\overline{\mathcal{S}}_c}=\mI-\mD_{\,\overline{\mathcal{S}}}\,,
\end{align}
which projects onto the complement of the expected sampling set, i.e., $\overline{\S}_c=\{i=1,\ldots,N \,|\, p_i=0\}$. Then, exploiting (\ref{exp_compl_oper}), signal reconstruction is possible if
$$\mU_{\F}^H\mD_{\,\overline{\mathcal{S}}}\mU_{\F}=\mI-\mU_{\F}^H\mD_{\,\overline{\mathcal{S}}_c}\mU_{\F}$$
is invertible, i.e., if condition
\begin{equation}\label{|DcB|<1}
   \boxed{\left\| \mD_{\,\overline{\mathcal{S}}_c}\mU_{\F}\right\|_2 < 1}
\end{equation}
holds true. As shown in \cite{TsitsveroEusipco15,tsitsvero2015signals}, condition (\ref{|DcB|<1}) is related to the localization properties of graph signals: It implies that there are no $\F$-bandlimited signals that are perfectly localized over the set $\oline{\S}_c$. Proceeding as in \cite{tsitsvero2015signals}, \cite{di2016least}, it is easy to show that condition (\ref{|DcB|<1}) is necessary and sufficient for signal reconstruction. We remark that, differently from previous works on sampling of graph signals, see, e.g.,
\cite{pesenson2008sampling,narang2013signal,chen2015discrete,tsitsvero2015signals,wang2014local,marquez2015,TsitsveroEusipco15}, condition (\ref{|DcB|<1}) now depends on the \textit{expected} sampling set. This relaxed condition is due to the iterative nature of the adaptive learning mechanism considered in this paper. As a consequence, the algorithm does not need to collect all the data necessary to reconstruct one-shot the graph signal at each iteration, but can learn acquiring the needed information over time. The only important thing required by condition (\ref{|DcB|<1}) is that a sufficiently large number of nodes collect data in \textit{expectation} (i.e., belong to the expected sampling set $\overline{\mathcal{S}}$). In the sequel, we introduce the proposed distributed algorithm.

\subsection{Adaptive Distributed Strategies}

In principle, the solution $\bs^o$ of problem (\ref{diffusion_LMS_problem}) can be computed as the vector that satisfies the linear system in (\ref{normal_equations}). Nevertheless, in many linear regression applications involving online processing of data, the moments in (\ref{normal_equations}) may be either unavailable or time-varying, and thus impossible to update continuously. For this reason, adaptive solutions relying on instantaneous information are usually adopted in order to avoid the need to know the signal statistics beforehand. Furthermore, the solution of (\ref{normal_equations}) would require to collect all the data $\{y_i[n]\}_{i:d_i[n]=1}$, for all $n$, in a single processing unit that performs the computation. In this paper, our emphasis is on distributed, adaptive solutions, where the nodes perform the reconstruction task via online in-network processing only exchanging data between neighbors. To this aim, diffusion techniques were proposed and studied in literature \cite{Cattivelli-Sayed,lopes2008diffusion,takahashi2010diffusion,fernandez2012novel,lopes2014towards}. In view of their robustness and adaptation properties, diffusion networks have been applied to solve a variety of learning tasks, such as, e.g., resource allocation problems \cite{di2013bio}, distributed optimization and learning \cite{Chen-Sayed}, sparse distributed estimation \cite{dilorenzo2013sparse,di2013distributed,dilorenzo2014diffusion},
robust system modeling  \cite{chouvardas2011adaptive}, and multi-task networks \cite{chen2014multitask,chen2015diffusion,chen2017multitask}.

In the sequel, we provide an alternative approach to derive diffusion adaptation strategies with respect to the seminal papers \cite{lopes2008diffusion,Cattivelli-Sayed}. The derivations will be instrumental to introduce the main assumptions that will be exploited in the mean-square analysis, which will be carried out in the next section. In particular, to ensure the diffusion of information over the entire network, the following is assumed:

\textit{Assumption 2 (Topology): The communication graph is symmetric and connected; i.e., there exists an undirected path connecting any two vertices of the network.} \qedsymbol

To derive distributed solution methods for problem (\ref{diffusion_LMS_problem}), let us introduce local copies $\{\bs_i\}_{i=1}^N$ of the global variable $\bs$, and recast problem (\ref{diffusion_LMS_problem}) in the following equivalent form:
\begin{align}\label{diffusion_LMS_problem2}
&\min_{\{\boldsymbol{s}_i\}_{i=1}^N} \;\; \sum_{i=1}^N \;\mathbb{E} \left|d_i[n]\left(y_i[n]-\bc_i^H\bs_i\right)\right|^2\\
%&\qquad \hbox{subject to} \quad \sum_{i=1}^N \sum_{j=1}^N a_{ij} \|\bs_j-\bs_i\|^2\leq 0 \nonumber
&\qquad \hbox{subject to} \quad \bs_i=\bs_j \quad \hbox{for all $i,j=1,\ldots,N$.} \nonumber
\end{align}
Under Assumption 2, it is possible to write the constraints in (\ref{diffusion_LMS_problem2}) in a compact manner, introducing the
disagreement constraint that enforces consensus among the local variables $\{\bs_i\}_{i=1}^N$ \cite{Barb-Sard-Dilo}. To this aim, let us denote with $\widetilde{\mA}=\{\widetilde{a}_{ij}\}$ the adjacency matrix of the communication graph among the nodes, such that $\widetilde{a}_{ij}>0$, if there is a communication link from node $j$ to node $i$, or $a_{ij}=0$, otherwise. Then, under Assumption 2, problem (\ref{diffusion_LMS_problem2}) [and (\ref{diffusion_LMS_problem})] can be rewritten in the following equivalent form:
\begin{align}\label{diffusion_LMS_problem3}
&\min_{\{\boldsymbol{s}_i\}_{i=1}^N} \;\; \sum_{i=1}^N \;\mathbb{E} \left|d_i[n]\left(y_i[n]-\bc_i^H\bs_i\right)\right|^2 \\
&\qquad \hbox{subject to} \quad \frac{1}{2}\sum_{i=1}^N \sum_{j=1}^N \widetilde{a}_{ij} \|\bs_j-\bs_i\|^2\leq 0. \nonumber
%&\qquad \hbox{subject to} \quad \bs_i=\bs_j \quad \hbox{for all $i,j=1,\ldots,N$.} \nonumber
\end{align}
The Lagrangian for problem (\ref{diffusion_LMS_problem3}) writes as:
\begin{align}\label{Lagrangian}
\mathcal{L}\left(\{\bs_i\}_{i=1}^N,\lambda^o\right)\,=\; & \sum_{i=1}^N \;\mathbb{E} \left|d_i[n]\left(y_i[n]-\bc_i^H\bs_i\right)\right|^2 \nonumber\\
&+\frac{\lambda^o}{2}\sum_{i=1}^N \sum_{j=1}^N \widetilde{a}_{ij} \|\bs_j-\bs_i\|^2
\end{align}
with $\lambda^o\geq0$ denoting the (optimal) Lagrange multiplier associated with the disagreement constraint. At this stage, we do not need to worry about the selection of the Lagrange multiplier $\lambda^o$, because it will be embedded into a set of coefficients that the designer can choose. Then, we proceed by minimizing the Lagrangian function in (\ref{Lagrangian}) by means of a steepest descent procedure. Thus, letting $\bs_i[n]$ be the instantaneous estimate of $\bs^o$ at node $i$, we obtain: \footnote{A factor of 2 multiplies (\ref{steepest_descent}) when the data are real. This factor was absorbed
into the step-sizes $\mu_i$ in (\ref{steepest_descent}).}
\begin{align}\label{steepest_descent}
\bs_i[n+1]\;&=\; \bs_i[n]-\mu_i \left[\nabla_{\boldsymbol{s}_i}\mathcal{L}\left(\{\bs_i[n]\}_{i=1}^N,\lambda^o\right)\right]^*\nonumber\\
\;&=\;\bs_i[n]+\mu_i \,\mathbb{E}\left\{d_i[n] \bc_i(y_i[n]-\bc_i^H\bs_i[n])\right\}\nonumber\\
&\textstyle\quad+\mu_i\lambda^o\sum_{j=1}^N \widetilde{a}_{ij} (\bs_j[n]-\bs_i[n])
\end{align}
for all $i=1,\ldots,N$, where $[\nabla(\cdot)]^*$ denotes the complex gradient operator, and $\mu_i>0$ are (sufficiently small) step-size coefficients. Now, using similar arguments as in \cite{Cattivelli-Sayed,Chen-Sayed,dilorenzo2013sparse,sayed2014adaptation}, we can accomplish the update (\ref{steepest_descent}) in two steps by generating an intermediate estimate $\boldsymbol{\psi}_i[n]$ as follows:
\begin{align}
&\boldsymbol{\psi}_i[n]=\bs_i[n]+\mu_i \,\mathbb{E}\left\{d_i[n] \bc_i(y_i[n]-\bc_i^H\bs_i[n])\right\}\label{adapt}\\
&\bs_i[n+1]=\boldsymbol{\psi}_i[n]+\mu_i\lambda^o\sum_{j=1}^N \widetilde{a}_{ij} (\boldsymbol{\psi}_j[n]-\boldsymbol{\psi}_i[n]) \label{combine}
\end{align}
where in (\ref{combine}) we have replaced the variables $\{\bs_i[n]\}_{i,n}$ with the intermediate estimates that are available at the nodes at time $n$, namely,  $\{\boldsymbol{\psi}_i[n]\}_{i,n}$. Such kind of substitutions are typically used to derive adaptive diffusion implementations, see, e.g., \cite{Cattivelli-Sayed}. Now, from (\ref{combine}), introducing the coefficients
\begin{align}\label{weights}
w_{ii}=1-\mu_i\lambda^o\sum_{j=1}^N \widetilde{a}_{ij}, \;\;\hbox{and} \;\; w_{ij}=\mu_i\lambda^o \widetilde{a}_{ij} \;\; \hbox{for $i\neq j$},
\end{align}
we obtain
\begin{align}
%&\boldsymbol{\psi}_i[n]=\bs_i[n]+\mu_i \,\mathbb{E}\left\{d_i[n] \bu_i^H(y_i[n]-\bu_i\bs_i[n])\right\}\label{adapt2}\\
\bs_i[n+1]=\sum_{j=1}^N w_{ij} \boldsymbol{\psi}_j[n] \label{combine2}
\end{align}
where the coefficients $\{w_{ij}\}$ are real, non-negative, weights matching the communication graph and satisfying:
\begin{align}\label{combination_coefficients}
w_{ij}=0 \;\; \hbox{for $j\notin \mathcal{N}_i$},\;\;\hbox{and}\;\; \mW\mathbf{1}=\mathbf{1},
\end{align}
where $\mW\in \mathbb{R}^{N\times N}$ is the matrix with individual entries $\{w_{ij}\}$, and $\mathcal{N}_i=\{j=1,\ldots,N\,|\,\widetilde{a}_{ij}>0\}\bigcup \{i\}$ is the extended neighborhood of node $i$, which comprises node $i$ itself. Recursion (\ref{adapt}) requires knowledge of the moments $\mathbb{E}\{d_i[n]y_i[n]\}$, which may be either unavailable or time-varying. An adaptive implementation can be obtained by replacing these moments by local instantaneous approximations, say, of the LMS type, i.e. $\mathbb{E}\{d_i[n]y_i[n]\}\approx d_i[n]y_i[n]$, for all $i,n$. The resulting algorithm is reported in Table 1, and will be termed as the Adapt-Then-Combine (ATC) diffusion strategy.
%\begin{algorithm}[h]
%\caption*{\textbf{Table 1: ATC diffusion for graph signal learning}}
%\vspace{.1cm}
%\textbf{Data:} $\bs_{i}[0]$ chosen at random for all $i$; $\{w_{ij}\}_{i,j}$ satisfying (\ref{combination_coefficients}); (sufficiently small) step-sizes $\mu_i>0$. Then, for each time $n\geq0$ and for each node $i$, repeat:
%\begin{align}\label{ATC diffusion}
%&\boldsymbol{\psi}_{i}[n]=\bs_{i}[n]+\mu_i d_i[n] \bc_i(y_i[n]-\bc_i^H\bs_i[n]) \nonumber\\
%&\hspace{4.6cm} \hbox{(adaptation step)} \\
%&\bs_i[n+1]=\sum_{j\in \mathcal{N}_i} w_{ij} \boldsymbol{\psi}_j[n] \hspace{.85cm} \hbox{(diffusion step)}\nonumber
%\end{align}
%\end{algorithm}
The first step in (\ref{ATC diffusion}) is an adaptation step, where the intermediate estimate $\boldsymbol{\psi}_{i}[n]$ is updated adopting the current observation taken by node $i$, i.e. $y_i[n]$.  The second step is a diffusion step where the intermediate estimates $\boldsymbol{\psi}_{j}[n]$, from the spatial neighbors $j\in \mathcal{N}_i$, are combined through the weighting coefficients $\{w_{ij}\}$. Finally, given the estimate $\bs_{i}[n]$ of the GFT at node $i$ and time $n$, the last step produces the estimate $x_i[n+1]$ of the graph signal value at node $i$ [cf. (\ref{lin_observation})]. We remark that by reversing the steps (\ref{adapt}) and (\ref{combine}) to implement (\ref{steepest_descent}), we would arrive at a similar but alternative strategy, known as the Combine-then-Adapt (CTA) diffusion strategy.
%We omit the description of the CTA implementation, and refer to \cite{Cattivelli-Sayed} for further details.
We continue our discussion by focusing on the ATC algorithm in (\ref{ATC diffusion}); similar analysis applies to CTA \cite{Cattivelli-Sayed}.
%Finally, given the estimate $\bs_{i}[n]$ of the GFT coefficients at node $i$ and time $n$, the estimate $x_i[n]$ of the graph signal value at node $i$, i.e. $x_i^o$ in (\ref{lin_observation}), can be computed locally as:
%\begin{align}\label{local_estimate}
%x_i[n]=\bc_i^H\bs_i[n], \quad \hbox{for all $i=1,\ldots,N$}.
%\end{align}

\textit{Remark 1:} The strategy (\ref{ATC diffusion}) allows each node in the network to estimate and track the (possibly time-varying) GFT of the graph signal $\bx^o$. From (\ref{ATC diffusion}), it is interesting to notice that, while sampling nodes (i.e., those such that $d_i[n]=1$) take and process the observations $y_i[n]$ of the graph signal, the role of the other nodes (i.e., those such that $d_i[n]=0$) is only to allow the propagation of information coming from neighbors through a diffusion mechanism that percolates over all the network. From a complexity point of view, at every iteration $n$, the strategy in (\ref{ATC diffusion}) requires that a node $i$ performs $O(3|\F|)$ computations, if $d_i[n]=1$, and $O(2|\F|)$ computations, if $d_i[n]=0$. Instead, from a communication point of view, each node in the network must transmit to its neighbors a vector composed of $|\F|$ (complex) values at every iteration $n$.

\begin{algorithm}[t]
\caption*{\textbf{Table 1: ATC diffusion for graph signal learning}}
\vspace{.1cm}
\textbf{Data:} $\bs_{i}[0]$ chosen at random for all $i$; $\{w_{ij}\}_{i,j}$ satisfying (\ref{combination_coefficients}); (sufficiently small) step-sizes $\mu_i>0$. Then, for each time $n\geq0$ and for each node $i$, repeat:
\begin{align}\label{ATC diffusion}
&\boldsymbol{\psi}_{i}[n]=\bs_{i}[n]+\mu_i d_i[n] \bc_i(y_i[n]-\bc_i^H\bs_i[n]) \nonumber\\
&\hspace{4.6cm} \hbox{(adaptation step)} \nonumber\\
&\bs_i[n+1]=\sum_{j\in \mathcal{N}_i} w_{ij} \boldsymbol{\psi}_j[n] \hspace{.85cm} \hbox{(diffusion step)}\\
&x_i[n+1]=\bc_i^H\bs_i[n+1]  \hspace{0.75cm} \hbox{(reconstruction step)}\nonumber
\end{align}
\end{algorithm}

In this work, we assume that processing and communication graphs have in general distinct topologies. We remark that both graphs play an important role in the proposed distributed processing strategy (\ref{ATC diffusion}). First, the processing graph determines the structure of the regression data $\bc_i$ used in the adaptation step of (\ref{ATC diffusion}). In fact, $\{\bc_i^H\}_i$ are the rows of the matrix $\mU_{\F}$, whose columns are the eigenvectors of the Laplacian matrix associated with the set of support frequencies $\F$. Then, the topology of the communication graph determines how the processed information is spread all over the network through the diffusion step in (\ref{ATC diffusion}). This illustrates how, when reconstructing graph signals in a distributed manner, we have to take into account both the processing and communication aspects of the problem. \qedsymbol

In the next section, we analyze the mean-square behavior of the proposed methods, enlightening the role played by the sampling strategy on the final performance.

%\begin{algorithm}[t]
%\caption*{\textbf{Table 2: CTA diffusion for Graph Signal Reconstruction}}
%\vspace{.1cm}
%\textbf{Data:} $\bs_{i}[0]$ chosen at random for all $i$; $\{w_{ij}\}_{i,j}$ satisfying (\ref{combination_coefficients}); (sufficiently small) step-sizes $\mu_i>0$. Then, for each time $n\geq0$ and for each node $i$, repeat:
%\begin{align}\label{CTA diffusion}
%&\boldsymbol{\chi}_{i}[n]=\displaystyle\sum_{j \in {\cal N}_i}w_{ij}\bs_{j}[n] \hspace{.6cm} \hbox{(combination step)}\nonumber\\
%&\hspace{4cm} \hbox{(adaptation step)} \\
%&\bs_{i}[n+1]=\boldsymbol{\chi}_{i}[n]+\mu_i d_i[n] \bu_i^H(y_i[n]-\bu_i\boldsymbol{\chi}_{i}[n]) \nonumber
%\end{align}
%\end{algorithm}

\section{Mean-Square Performance Analysis}

In this section, we analyze the performance of the ATC strategy in (\ref{ATC diffusion}) in terms of its mean-square behavior.
We remark that the analysis carried out in this section differs from classical derivations used for diffusion adaptation algorithms, see, e.g., \cite{sayed2014adaptation}. First of all, the observation model in (\ref{lin_observation}) is different from classical models generally adopted in the adaptive filtering literature, see, e.g. \cite{sayed2011adaptive}. Also, due to the sampling operation and the presence of deterministic regressors [cf. (\ref{lin_observation})], each node cannot reconstruct the signal using only its own data (i.e., using stand-alone LMS adaptation), and must necessarily cooperate with other nodes in order exploit information coming from other parts of the network. These issues require the development of a dedicated (non-trivial) analysis (see, e.g., Theorem 1 and the Appendix) to prove the mean-square stability of the proposed algorithm. %As a consequence, the (mean-square) convergence properties of the algorithm will depend on some global conditions on the step-sizes $\{\mu_i\}$ and the expected sampling set $\overline{\mathcal{S}}$.}

From now on, we view the estimates $\bs_i[n]$ as realizations of a random process and analyze the performance of the ATC diffusion algorithm in terms of its mean-square behavior. To do so, we introduce the error quantities
$$\bee_i[n]=\bs_i[n]-\bs^o,\quad i=1,\ldots,N,$$ and the network vector
\begin{eqnarray}\label{err_vectors}
\bee[n]={\rm col}\{\bee_{1}[n], \ldots,\bee_{N}[n]\}.
\end{eqnarray}
We also introduce the block diagonal matrix
\begin{eqnarray}\label{step_matrix}
\mM={\rm diag}\{\mu_1\mI_{|\F|},\ldots,\mu_N\mI_{|\F|}\},
\end{eqnarray}
the extended block weighting matrix
\begin{eqnarray}\label{combination_matrices}
\widehat{\mW}=\mW\otimes \mI_{|\F|},
\end{eqnarray}
where $\otimes$ denotes the Kronecker product operation, and the extended sampling operator
\begin{eqnarray}\label{sampling_operator}
\widehat{\mD}[n]={\rm diag}\left\{d_1[n]\mI_{|\F|},\ldots,d_N[n]\mI_{|\F|}\right\}.
\end{eqnarray}
We further introduce the block quantities:
\begin{align}
\mQ&={\rm diag}\left\{\bc_{1}\bc_{1}^H,\ldots,\bc_{N}\bc_{N}^H\right\}, \label{perf_matrices}\\
\bg[n]&={\rm col}\{\bc_{1}v_1[n],\ldots,\bc_{N}v_N[n]\}.\label{perf_matrices2}
\end{align}
Then, exploiting (\ref{err_vectors})-(\ref{perf_matrices2}), we conclude from (\ref{ATC diffusion}) that the following relation holds for the error vector:
\begin{align}\label{compact_Diffusion}
\bee[n+1]=\widehat{\mW}\left(\mI-\mM\widehat{\mD}[n]\mQ\right)\bee[n]+\widehat{\mW}\mM\widehat{\mD}[n]\bg[n].
\end{align}
This relation tells us how the network error vector evolves over time. As we can notice from (\ref{compact_Diffusion}), the sampling strategy affects the recursion in two ways: (a) it modifies the iteration matrix $\widehat{\mW}(\mI-\mM\widehat{\mD}[n]\mQ)$ of the error; (b) it selects the noise contribution $\widehat{\mD}[n]\bg[n]$ only from a subset of nodes at time $n$. Relation (\ref{compact_Diffusion}) will be the launching point for the mean-square analysis carried out in the sequel. Before moving forward, we introduce an independence assumption on the sampling strategy, and a small step-sizes assumption.

\vspace{.1cm}
{\textit{Assumption 3 (Independent sampling)}:  The sampling process $\{d_i[t]\}$ is temporally and spatially independent, for all $i=1,\ldots,N$ and $t\leq n$.} \qedsymbol

\vspace{.1cm}
{\textit{Assumption 4 (Small step-sizes):  The step-sizes $\{\mu_i\}$ are sufficiently small so that terms that depend on higher-order powers of $\{\mu_i\}$ can be neglected.}}{\qedsymbol}

\vspace{.1cm}
We now proceed by illustrating the mean-square stability and steady-state performance of the algorithm in (\ref{ATC diffusion}).

\subsection{Mean-Square Stability}
We now examine the behavior of the mean-square deviation $\mathbb{E}\|\bee_i[n]\|^2$ for any of the nodes in the graph. Following energy conservation arguments \cite{Cattivelli-Sayed,sayed2014adaptation}, we can establish the following variance relation:
\begin{align}\label{weighted_norm_expanded}
\mathbb{E}\|\bee[n+1]\|^2_{\boldsymbol{\Sigma}}=& \; \mathbb{E}\|\bee[n]\|^2_{\boldsymbol{\Sigma}'} \nonumber\\
&\hspace{-1.5cm}+\mathbb{E}\left[\bg[n]^H\widehat{\mD}[n]\mM\widehat{\mW}^T\boldsymbol{\Sigma} \widehat{\mW}\mM\widehat{\mD}[n]\bg[n]\right]
\end{align}
where $\boldsymbol{\Sigma}$ is any Hermitian nonnegative-definite matrix that we are free to choose, and
\begin{eqnarray}\label{Sigma'}
\boldsymbol{\Sigma}'=\mathbb{E} \left(\mI-\mQ\widehat{\mD}[n]\mM\right)\widehat{\mW}^T \boldsymbol{\Sigma}\widehat{\mW}\left(\mI-\mM\widehat{\mD}[n]\mQ\right).\label{Sigma_primo}
\end{eqnarray}
Moreover, setting
\begin{eqnarray}\label{matrixG}
\mG=\mathbb{E}\left[\bg[n]\bg[n]^H\right]={\rm diag}\left\{\sigma_{1}^2 \bc_{1}\bc_{1}^H,\ldots,\sigma_{N}^2 \bc_{N}\bc_{N}^H\right\},
\end{eqnarray}
we can rewrite (\ref{weighted_norm_expanded}) in the form
\begin{eqnarray}\label{weighted_norm2}
\mathbb{E}\|\bee[n+1]\|^2_{\boldsymbol{\Sigma}}=\mathbb{E}\|\bee[n]\|^2_{\boldsymbol{\Sigma}'}
+{\rm Tr}\left(\boldsymbol{\Sigma} \widehat{\mW}\mM\widehat{\mP}\mG\mM\widehat{\mW}^T\right)
\end{eqnarray}
where ${\rm Tr}(\cdot)$ denotes the trace operator,
$$\widehat{\mP}=\mathbb{E}\,\left\{\widehat{\mD}[n]\right\}=\mP\otimes \mI_{|\F|},$$
and we have exploited the relation [cf. (\ref{sampling_operator}), (\ref{matrixG}), and Assumption 3]
$$\mathbb{E}\left\{\widehat{\mD}[n]\bg[n]\bg[n]^H\widehat{\mD}[n]\right\}=\mathbb{E}\left\{\widehat{\mD}[n]\bg[n]\bg[n]^H\right\}=\widehat{\mP}\mG.$$
Let $\boldsymbol{\sigma}={\rm vec}(\boldsymbol{\Sigma})$ and $\boldsymbol{\sigma}'=\mbox{\rm vec}(\boldsymbol{\Sigma}')$,
where the ${\rm vec}(\cdot)$ notation stacks the columns of $\boldsymbol{\Sigma}$ on top of each other and ${\rm vec}^{-1}(\cdot)$ is the inverse operation. We will use interchangeably the notation $\|\bee\|^2_{\boldsymbol{\sigma}}$ and $\|\bee\|^2_{\boldsymbol{\Sigma}}$ to denote the same quantity $\bee^H\boldsymbol{\Sigma}\bee$.
Using the Kronecker product property ${\rm vec}(\mA\boldsymbol{\Sigma} \mC)=(\mC^T\otimes \mA){\rm vec}(\boldsymbol{\Sigma})$,
we can vectorize both sides of (\ref{Sigma'}) and conclude that (\ref{Sigma'}) can be replaced by the simpler linear vector relation: $\boldsymbol{\sigma}'={\rm vec}(\boldsymbol{\Sigma}')=\mH\boldsymbol{\sigma}$, where $\mH$ is the $N^2|\F|^2\times N^2|\F|^2$ matrix:
\begin{align}\label{matrix_H}
\mH &= \mathbb{E}\left\{\left(\mI-\mQ^T\widehat{\mD}[n]\mM\right)\widehat{\mW}^T\otimes \left(\mI-\mQ\widehat{\mD}[n]\mM\right)\widehat{\mW}^T   \right\} \nonumber\\
&= (\mI\otimes\mI)\Big(\mI-\mI\otimes \mQ\widehat{\mP}\mM- \mQ^T\widehat{\mP}\mM\otimes\mI  \nonumber\\
&\hspace{.5cm} + \mathbb{E}\left\{\mQ^T\widehat{\mD}[n]\mM \otimes \mQ\widehat{\mD}[n]\mM \right\} \Big) \big(\mW^T\otimes\mW^T\big)
\end{align}
The last term in (\ref{matrix_H}) can be computed in closed form. In particular, from (\ref{step_matrix}), (\ref{sampling_operator}), and (\ref{perf_matrices}), it is easy to see how the term $\mQ\widehat{\mD}[n]\mM$ (and $\mQ^T\widehat{\mD}[n]\mM$) in (\ref{matrix_H}) has a block diagonal structure, which can be recast as:
\begin{align}\label{Block}
&\mQ\widehat{\mD}[n]\mM = \sum_{i=1}^N \mu_i d_i[n] \mC_i,
\end{align}
where $\mC_i=\bc_i\bc_i^H\otimes\mR_i$, and $\mR_i={\rm diag}(\br_i)$, with $\br_i$ denoting the $i$-th canonical vector. Thus, exploiting (\ref{Block}), we obtain
\begin{align}\label{Block2}
&\mathbb{E}\left\{\mQ^T\widehat{\mD}[n]\mM \otimes \mQ\widehat{\mD}[n]\mM \right\}\nonumber\\
&\qquad\qquad= \sum_{i=1}^N \sum_{j=1}^N \mu_i\mu_j m^{(2)}_{i,j}\mC_i^T\otimes\mC_j
\end{align}
where, exploiting Assumption 3, we have
\begin{align}\label{Block3}
\quad m^{(2)}_{i,j}=\mathbb{E}\{d_i[n]d_j[n]\}=    \begin{cases}
      p_i, & \text{if}\;\; i=j; \\
      p_ip_j, & \text{if}\;\; i\neq j.
    \end{cases}
\end{align}
Substituting (\ref{Block2}) in (\ref{matrix_H}), we obtain the final expression:
\begin{align}\label{matrix_H2}
\mH &= (\mI\otimes\mI)\bigg(\mI-\mI\otimes \mQ\widehat{\mP}\mM- \mQ^T\widehat{\mP}\mM\otimes\mI  \nonumber\\
&\hspace{.5cm} + \sum_{i=1}^N \sum_{j=1}^N \mu_i\mu_j m^{(2)}_{i,j}\mC_i^T\otimes\mC_j \bigg) \big(\mW^T\otimes\mW^T\big).
\end{align}

Now, using the property ${\rm Tr}(\boldsymbol{\Sigma} \mX)={\rm vec}(\mX^T)^T\boldsymbol{\sigma}$, we can rewrite (\ref{weighted_norm2}) as follows:
\begin{equation}\label{weighted_norm3}
\mathbb{E}\|\bee[n+1]\|^2_{\boldsymbol{\sigma}}=\mathbb{E}\|\bee[n]\|^2_{\mathbf{H}\boldsymbol{\sigma}}+{\rm vec}\left(\widehat{\mW}\mM\widehat{\mP}\mG\mM\widehat{\mW}^T\right)^T\boldsymbol{\sigma}
\end{equation}
The following theorem guarantees the asymptotic mean-square stability (i.e., stability in the mean and mean-square sense) of the diffusion strategy (\ref{ATC diffusion}).

\vspace{.1cm}
\noindent {\textit{Theorem 1 (Mean-square stability)}} Assume model (\ref{lin_observation}), condition (\ref{|DcB|<1}), Assumptions 2, 3, and 4  hold. Then, for any initial condition and choice of the matrices $\mW$ satisfying (\ref{combination_coefficients}) and $\mathbf{1}^T\mW=\mathbf{1}^T$, the algorithm (\ref{ATC diffusion}) is mean-square stable.

\vspace{.1cm}
\begin{proof}
Let $\br={\rm vec}\left(\widehat{\mW}\mM\widehat{\mP}\mG\mM\widehat{\mW}^T\right)$. From (\ref{weighted_norm3}), we get
\begin{equation}\label{weighted_norm_last}
\mathbb{E}\|\bee[n]\|^2_{\boldsymbol{\sigma}}=\mathbb{E}\|\bee[0]\|^2_{\mathbf{H}^n\boldsymbol{\sigma}}+\br^T\sum_{l=0}^{n-1}\mH^l\boldsymbol{\sigma}
\end{equation}
where $\mathbb{E}\|\bee[0]\|^2$ is the initial condition. We first note that if $\mH$ is stable, $\mH^n \rightarrow \mathbf{0}$ as $n\rightarrow\infty$. In this way, the first term on the RHS of (\ref{weighted_norm_last}) vanishes asymptotically. At the same time, the convergence of the second term on the RHS of (\ref{weighted_norm_last}) depends only on the geometric series of matrices $\sum_{l=0}^{\infty}\mH^l$, which is known to be convergent to a finite value if the matrix $\mH$ is a stable matrix \cite{horn2012matrix}. Thus, the stability of matrix $\mH$ is a sufficient condition for the convergence of the mean-square recursion $\mathbb{E}\|\bee[n]\|^2_{\boldsymbol{\sigma}}$ in (\ref{weighted_norm_last}) to a steady-state value. \\
\indent To verify the stability of $\mH$, we use the following approximation, which is accurate under Assumption 4, see, e.g., \cite{Cattivelli-Sayed,Chen-Sayed,dilorenzo2013sparse}.
%note that the step sizes influence (\ref{matrix_H}) through the matrix $\mM$. Since in virtue of Assumption 4 the step-sizes are sufficiently small, terms that depend on higher-order powers of the step-sizes have a small effect.
Then, we approximate (\ref{matrix_H}) as:
\footnote{It is immediate to see that (\ref{approx_matrix_F}) can be obtained from (\ref{matrix_H2}) by replacing the term $\mathbb{E}\left\{\mQ^T\widehat{\mD}[n]\mM \otimes \mQ\widehat{\mD}[n]\mM\right\}$ with $\mQ^T\widehat{\mP}\mM \otimes \mQ\widehat{\mP}\mM$. This step coincides with substituting the terms $p_i$ in (\ref{Block2})-(\ref{Block3}) with $p_i^2$, for all $i=1,\ldots,N$. Such approximation appears in (\ref{approx_matrix_F}) only in the term $\mQ^T\widehat{\mP}\mM \otimes \mQ\widehat{\mP}\mM=O(\mu_{\max}^2)$ and, consequently, under Assumption 4 it is assumed to produce a negligible deviation from (\ref{matrix_H2}).}
\begin{align}\label{approx_matrix_F}
\mH &\;\approx\;  (\mI\otimes\mI)\Big(\mI-\mI\otimes \mQ\widehat{\mP}\mM- \mQ^T\widehat{\mP}\mM\otimes\mI  \nonumber\\
&  + \mQ^T\widehat{\mP}\mM \otimes \mQ\widehat{\mP}\mM \Big) (\mW^T\otimes\mW^T)\;=\; \mB^T\otimes \mB^H
\end{align}
with $\mB$ given by
\begin{align}\label{matrixB}
\mB=\widehat{\mW}\left(\mI-\mM\widehat{\mP}\mQ\right).
\end{align}
Thus, from (\ref{approx_matrix_F}), exploiting the properties of the Kronecker product, we deduce that matrix $\mH$ in (\ref{matrix_H}) is stable if matrix $\mB$ in (\ref{matrixB}) is also stable. Under the assumptions of Theorem 1, in the Appendix, we provide the proof of the stability of matrix $\mB$ in (\ref{matrixB}). This concludes the proof of Theorem 1.
\end{proof}

\vspace{.1cm}
\noindent {\textit{Remark 2:}} The assumptions used in Theorem 1 are \textit{sufficient} conditions for graph signal reconstruction using the ATC diffusion algorithm in (\ref{ATC diffusion}).  In particular, (\ref{|DcB|<1}) requires that the network collects samples from a sufficiently large number of nodes on average, and guarantees the existence of a \textit{unique} solution of the normal equations in (\ref{normal_equations}). Furthermore, (\ref{|DcB|<1}) and Assumption 4 are also instrumental to prove the stability of matrix $\mB$ in (\ref{matrixB}) [and of $\mH$ in (\ref{matrix_H})] and, consequently, the stability in the mean and mean-square sense of the diffusion algorithm in (\ref{ATC diffusion}) (see the Appendix). \qedsymbol

\vspace{.1cm}
{\textit{Remark 3:}} In Theorem 1, we require the matrix $\mW$ to be doubly stochastic. Note that, from the definition of weights $\{w_{ij}\}$ in (\ref{weights}), under assumption 2, this further condition would imply that the step-sizes $\mu_i$ must be chosen constant for all $i$. However, as a consequence of Theorem 1, our strategy works for \textit{any} choice of doubly stochastic matrices $\mW$, without imposing the constraint that the step-sizes must be chosen constant for all $i$. Several possible combination rules have been proposed in the literature, such as the Laplacian or the Metropolis-Hastings weights, see, e.g. \cite{Barb-Sard-Dilo}, \cite{xiao2007distributed}, \cite{Cattivelli-Sayed}.\qedsymbol

\subsection{Steady-State Performance}

Taking the limit as $n \rightarrow \infty$ (assuming convergence conditions are satisfied), we deduce from (\ref{weighted_norm3}) that:
\begin{equation}\label{variance_relation2}
\displaystyle \lim_{n\rightarrow\infty}\mathbb{E}\|\bee[n]\|^2_{(\mathbf{I}-\mathbf{H})\boldsymbol{\sigma}}={\rm vec}\left(\widehat{\mW}\mM\widehat{\mP}\mG\mM\widehat{\mW}^T\right)^T\boldsymbol{\sigma}.
\end{equation}
Expression (\ref{variance_relation2}) is a useful result: it allows us to derive several performance metrics through the proper selection of the free weighting parameter $\boldsymbol{\sigma}$ (or $\boldsymbol{\Sigma}$), as was done in \cite{Cattivelli-Sayed}. For example, the Mean-Square Deviation (MSD) for any node $i$ is defined as the steady-state value $\mathbb{E}|\widetilde{x}_i[n]|^2$, as $n\rightarrow\infty$, where $\widetilde{x}_i[n]=x_i[n]-x_i^o[n]$, for all $i=1,\ldots,N$, with $x_i[n]$ defined in (\ref{ATC diffusion}). From (\ref{ATC diffusion}), this value can be obtained by computing $\lim_{n\rightarrow\infty}\mathbb{E}\|\bee[n]\|^2_{\mathbf{T}_i}$, with a block weighting matrix $\mT_i=\mR_i\otimes\bc_i\bc_i^H$, where $\mR_i={\rm diag}(\br_i)$, with $\br_i$ denoting the $i$-th canonical vector. Then, from (\ref{variance_relation2}), the MSD at node $i$ can be obtained as:
\begin{align}\label{MSDk}
{\rm MSD}_i&=\displaystyle \lim_{n\rightarrow\infty}\mathbb{E}\,|\widetilde{x}_i[n]|^2=\displaystyle \lim_{n\rightarrow\infty}\mathbb{E}\|\bee[n]\|^2_{\mathbf{R}_i\otimes\bc_i\bc_i} \\
&\hspace{-.6cm}={\rm vec}\left(\widehat{\mW}\mM\widehat{\mP}\mG\mM\widehat{\mW}^T\right)^T(\mI-\mH)^{-1}{\rm vec}\left(\mR_i\otimes\bc_i\bc_i^H\right). \nonumber
\end{align}
Finally, letting $\widetilde{\bx}[n]=\{\widetilde{x}_i[n]\}_{i=1}^N$, from (\ref{MSDk}), the network ${\rm MSD}$ is given by:
\begin{align}\label{MSD_net}
{\rm MSD}&=\displaystyle \lim_{n\rightarrow\infty}\mathbb{E}\|\widetilde{\bx}[n]\|^2  \nonumber\\%=\displaystyle \nonumber \lim_{n\rightarrow\infty}\mathbb{E}\|\widetilde{\bs}[n]\|^2 \\
&={\rm vec}\left(\widehat{\mW}\mM\widehat{\mP}\mG\mM\widehat{\mW}^T\right)^T(\mI-\mH)^{-1}\bq,
\end{align}
where
$\bq={\rm vec}\left(\sum_{i=1}^N\mR_i\otimes\bc_i\bc_i^H\right)={\rm vec}\left(\mQ\right)$ [cf. (\ref{perf_matrices})]. In the sequel, we will confirm the validity of these theoretical expressions by comparing them with numerical results.

\section{Distributed Graph Sampling Strategies}

As illustrated in the previous sections, the properties of the proposed distributed algorithm in (\ref{ATC diffusion}) for graph signal reconstruction strongly depend on the expected sampling set $\overline{\mathcal{S}}$. Thus, building on the results obtained in Sec. IV, it is fundamental to devise (possibly distributed) strategies that design the set $\overline{\mathcal{S}}$, with the aim of reducing the computational/memory burden while still guaranteing provable theoretical performance. To this purpose, in this section we propose a distributed method that iteratively selects vertices from the graph in order to build an expected sampling set $\overline{\mathcal{S}}$ that enables reconstruction with a limited number of nodes, while ensuring guaranteed performance of the learning task.

To select the best sampling strategy, one should optimize some (global) performance criterion, e.g. the MSD in (\ref{MSD_net}), with respect to the expected sampling set $\overline{\S}$, or, equivalently, the weighted vertex limiting operator $\widehat{\mP}$. However, the solution of such problem would require global information about the entire graph to be collected into a single processing unit. To favor distributed implementations, we propose to consider a different (but related) performance metric for the selection of the sampling set, which comes directly from the solution of the normal equations in (\ref{normal_equations}). In particular, to allow reconstruction of the graph signal, a good sampling strategy should select a sufficiently large number of vertices $i\in \V$ to favor the invertibility of the matrix in (\ref{cond}). In the sequel, we assume that the probabilities $\{p_i\}_{i=1}^N$ are given, either because they are known apriori or can be estimated locally at each node. In this context, the design of the sampling probabilities $\{p_i\}_{i=1}^N$ is an important task, which will be tackled in a future work.

Let us then consider the general selection problem:
\begin{align}\label{sampling_problem}
&\S^*=\,\arg\max_{\overline{\S}}\;h\left(\overline{\S}\right)=f\left(\sum_{i\in \overline{\S}} \frac{p_i}{1+\sigma_i^2}\, \bc_i\bc_i^H\right) \\
&\hspace{1.2cm} \hbox{subject to} \quad|{\overline{\S}}|=M \nonumber
\end{align}
where ${\overline{\S}}$ is the expected sampling set; $M$ is the given number of vertices samples to be selected;
the weighting terms $p_i/(1+\sigma_i^2)$ take into account (possibly) heterogeneous sampling and noise conditions at each node;
%$\bp=[p_1,\ldots,p_N]^T\in\mathbf{R}^N$ is the vector collecting the sampling probabilities at each vertex;
and $f(\cdot):\mathbb{C}^{|\F|\times|\F|}\rightarrow \mathbb{R}$ is a function that measures the degree of invertibility of the matrix in its argument, e.g., the (logarithm of) pseudo-determinant, as proposed in \cite{tsitsvero2015signals}, \cite{di2016least}, \cite{chepuri2016subsampling}, or the minimum eigenvalue, as proposed in \cite{chen2015discrete}. As an example, taking $f(\cdot)$ as the (logarithm of) pseudo-determinant function, the solution of problem (\ref{sampling_problem}) aims at selecting $M$ rows $\bc_i^H$ of matrix $\mU_\F$, properly weighted by the terms $\sqrt{p_i/(1+\sigma_i^2)}$, such that the volume of the parallelepiped built by these vectors is maximized. Thus, intuitively, the method will tend to select vertices with: (a) large sampling probabilities $p_i$'s; (b) low noise variances  $\sigma_i^2$'s; and (c) such that their corresponding regression vectors $\bc_i$'s are large in magnitude and as orthogonal as possible. However, since the formulation in (\ref{sampling_problem}) translates inevitably into a selection problem, whose solution in general requires an exhaustive search over all the possible combinations, the complexity of such procedure becomes intractable also for graph signals of moderate dimensions. To cope with these issues, in the sequel we will provide an efficient, albeit sub-optimal, greedy strategy that tackles the problem of selecting the (expected) sampling set in a distributed fashion.

The greedy approach is described in table 2. The simple idea underlying the proposed approach is to iteratively add to the sampling set those vertices of the graph that lead to the largest increment of the performance metric $h\left(\overline{\S}\right)$ in (\ref{sampling_problem}). In particular, the implementation of the distributed algorithm in Table 2 proceeds as follows. Given the current instance of the (expected) sampling set $\overline{\S}$, at Step 1, each node $j\notin\overline{\S}$ evaluates locally the value of the objective function $h\left(\overline{\S}\cup j\right)$ that the network would achieve if node $j$ was added to $\overline{\S}$. Then, in step 2, the network finds the maximum among the local values computed at the previous step. This task can be easily obtained with a distributed iterative procedure as, e.g., a maximum consensus algorithm \cite{olfati2004consensus}, which is guaranteed to converge in a number of iterations less than equal to $\mathcal{D}$, with $\mathcal{D}$ denoting the diameter of the network. A node $j\notin\overline{\S}$ can then understand if it is the one that has achieved the maximum by simply comparing the value $h\left(\overline{\S}\cup j\right)$ computed at step 1, with the result of the distributed procedure in Step 2. The node $s^*$, which has achieved the maximum value at step 2, is then added to the expected sampling set. Finally, the weighted regression vector associated to the selected node, i.e. $\sqrt{p_{s^*}/(1+\sigma_{s^*}^2)}\bc_{s^*}$, is diffused over the network through a flooding process, which terminates in a number of iterations less than or equal to $\mathcal{D}$. This allows each node not belonging to the sampling set to evaluate step 1 of the algorithm at the next round. This procedure continues until the network has added $M$ nodes to the expected sampling set.

\begin{algorithm}[t]

\vspace{.1cm}
$\textit{Input Data}:$ $M$, the number of samples. $\overline{\S}\equiv \emptyset$.

$\textit{Output Data}:$ $\overline{\S}$, the expected sampling set. \smallskip

$\textit{Function}:$

\smallskip
\hspace{.2cm}while $|\overline{\S}|<M$ \smallskip

\begin{enumerate}
  \item Each node $j$ computes locally $h\left(\overline{\S}\cup j\right)$, for all $j\notin \overline{\S}$;
  \item Distributed selection of the maximum: find $$s^*=\arg \max_{j\notin \overline{\S}} \;\; h\left(\overline{\S}\cup j\right)$$
  \item $\overline{\S} \leftarrow \overline{\S} \cup \{s^*\}$; \smallskip
  \item Diffusion of $\displaystyle\sqrt{\frac{p_{s^*}}{1+\sigma_{s^*}^2}}\,\bc_{s^*}$ over the network; \smallskip
\end{enumerate}

%\hspace{2.1cm} \textit{S.1:} The network runs $N-|\S|$ consensus rounds to compute $h\left(\bd_{\S\cup j}\right)$, for all $j\notin \S$;
%
%\hspace{2.1cm} \textit{S.2:} Each node performs locally $\displaystyle s^*=\arg \max_{j\notin \S} \;\; h\left(\bd_{\S\cup j}\right)$;
%
%\hspace{2.1cm} \textit{S.3:} If $j=s^*$, node $j$ is added to the sampling set, i.e. $\S \leftarrow \S \cup \{j\}$;

\hspace{.2cm}end \vspace{.1cm}

\protect\caption*{\label{alg:Greedy}\textbf{Table 2: Distributed Graph Sampling Strategy}}
\end{algorithm}

In principle, there is no insurance that the selection path followed by the algorithm in Table 2 is the best one. In general, the performance of the proposed distributed strategy will be suboptimal with respect to an exhaustive search procedure over all the possible combinations. Nevertheless, selecting the function $h\left(\overline{\S}\right)$ in (\ref{sampling_problem})
%=\log \det\left(\sum_{i\in \overline{\S}} \frac{p_i}{1+\sigma_i^2}\, \bc_i\bc_i^H\right)$
as the logarithm of the pseudo determinant, it is possible to prove that $h\left(\overline{\S}\right)$ is a monotone sub-modular function, and that greedy selection strategies (e.g., Table 2) achieve performance within $1-1/e$ of the optimal combinatorial solution \cite{chepuri2015sparsity,shamaiah2010greedy}.
%runs $N-|\S|$ average consensus rounds. This allows each node to know locally the values of the metric $h\left(\bd_{\S\cup j}\right)$, for all $j\notin \S$. \footnote{Note that there are also implementations of consensus algorithms where consensus is achieved in finite-time, see, e.g. \cite{wang2010finite} and references therein.} At this point, the maximum operation in Table 3 can be performed locally, and the node that attains the maximum value is added to the sampling set $\S$.
From a communication point of view, in the worst case, the procedure in Table 2 requires that each node exchanges $M\mathcal{D}(1+2|\F|)$ scalar values to accomplish the distributed task of sampling set selection. This procedure can be run offline once for all during the initialization phase of the network, when the set of sampling nodes must be decided. In the case of time-varying scenarios, e.g. switching graph topologies, link failures, time-varying spectral properties of the graph signal, the procedure should be repeated periodically in order to cope with such dynamicity. Of course, the procedure might result unpractical in the case of large, rapidly time-varying graphs. In such a case, future investigations are needed for practical and efficient implementations of distributed adaptive graph sampling strategies.

In the sequel, we will illustrate numerical results assessing the performance achieved by the proposed sampling strategies.

\section{Numerical Results}

In this section, we illustrate some numerical simulations aimed at assessing the performance of the proposed strategy for distributed learning of signals defined over graphs. First, we will illustrate the convergence properties of the proposed algorithm in absence of observation noise. Second, we will confirm the theoretical results in (\ref{weighted_norm3}) and (\ref{MSDk})-(\ref{MSD_net}), which quantify the transient behavior and steady-state performance of the algorithm. Third, we will illustrate how the choice of the sampling strategy (see, e.g., Table 2) affects the performance of the proposed algorithm. Fourth, we will evaluate the tracking capabilities of the proposed technique, considering the presence of stochastic processes evolving over the graph. Finally, we apply the proposed strategy to estimate and track the spatial distribution of electromagnetic power in the context of cognitive radio networks.

\begin{figure}[t]
\centering
\includegraphics[width=7.5cm]{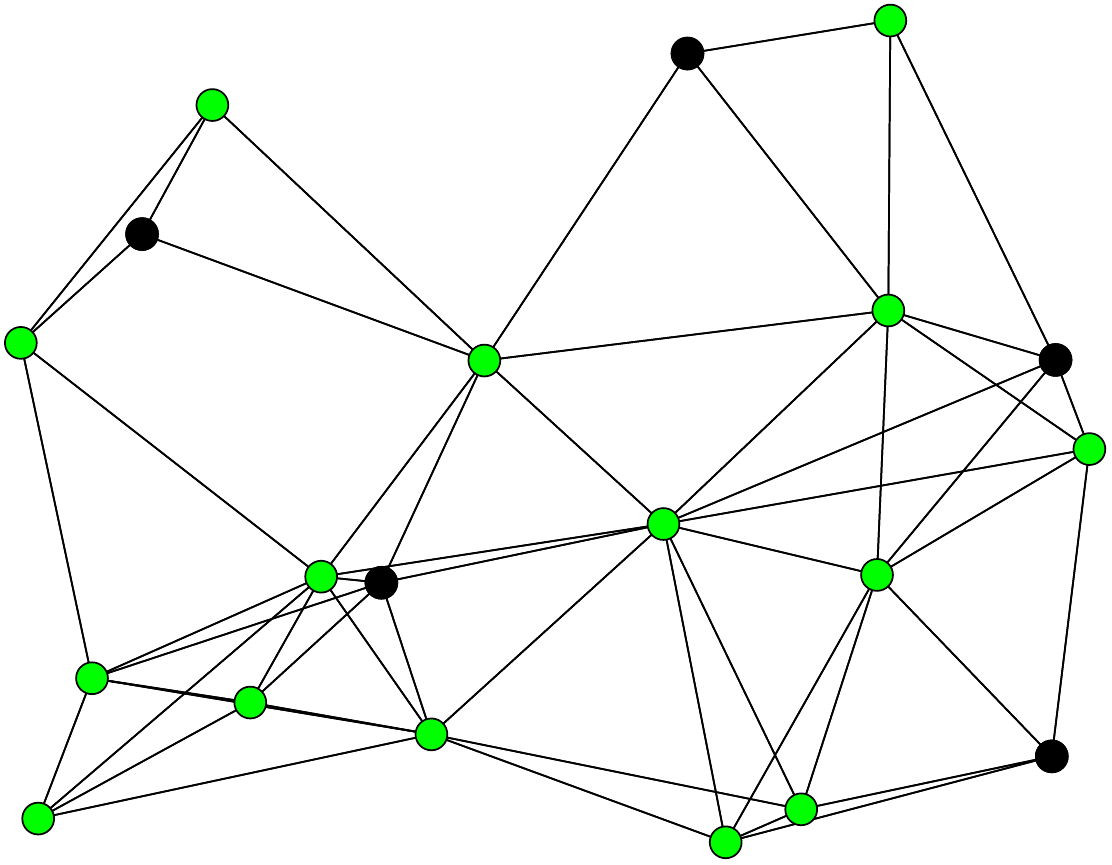}
  \caption{Network graph, and sampling set (black nodes).}\label{fig:Network}
\end{figure}

\subsubsection{Convergence in the noise-free case}

\begin{figure}[t]
\centering
\includegraphics[width=8.5cm]{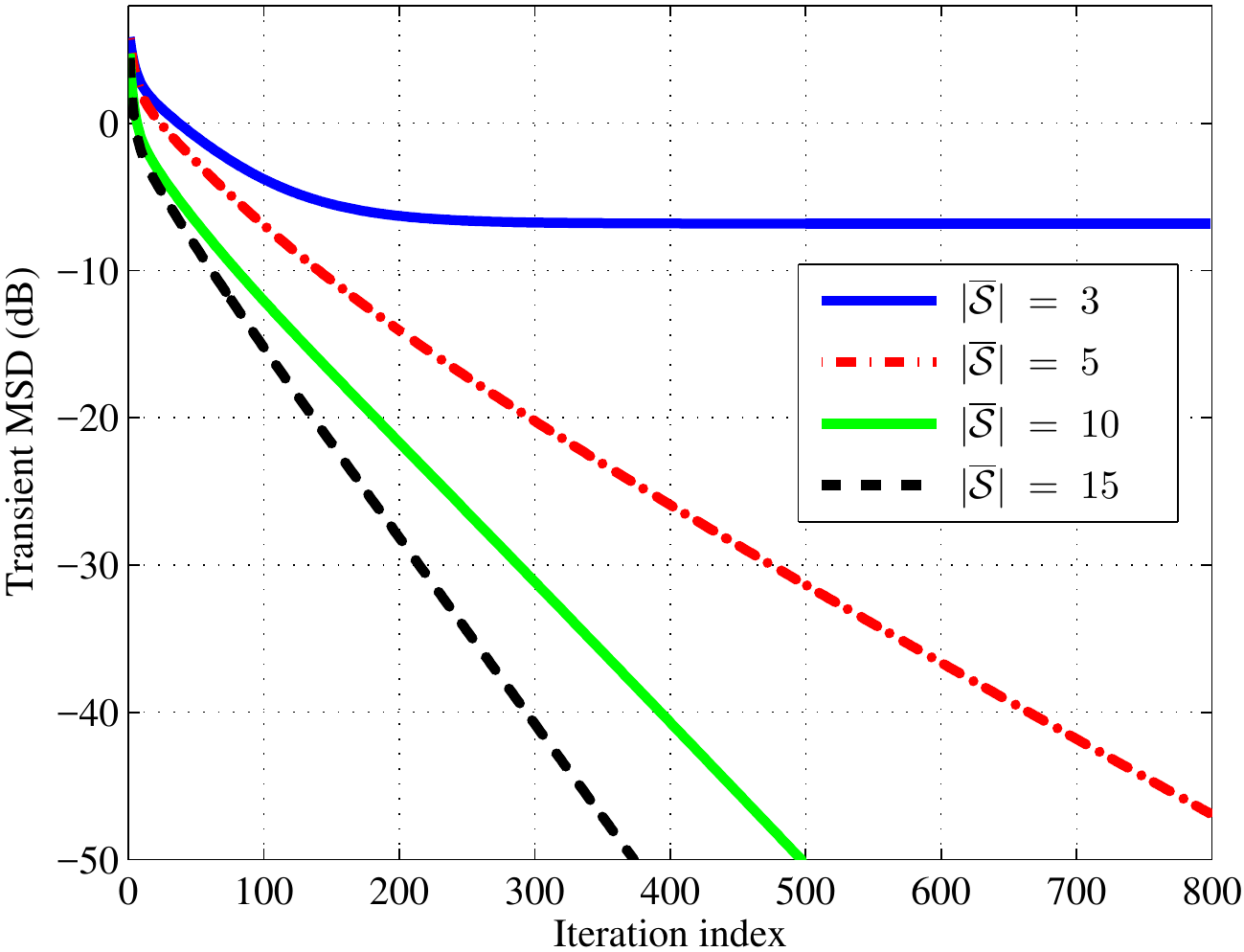}
  \caption{Convergence behavior: Transient MSD in the noise-free case, considering static sampling. $|\mathcal{F}|=5$.}\label{convergence}
\end{figure}

Let us consider a network composed of $N=20$ nodes, whose topology (for both processing and communication tasks) is depicted in Fig. \ref{fig:Network}. We generate a graph signal from (\ref{x=Us}) having a spectral content limited to the first five eigenvectors of the Laplacian matrix of the graph in Fig. \ref{fig:Network}. Thus, the signal bandwidth is equal to $|\mathcal{F}|=5$. For simplicity, we use the graph illustrated in Fig. \ref{fig:Network} for both communication and processing tasks. To illustrate the perfect reconstruction capabilities of the proposed method in absence of noise, in this simulation we set $v_i[n]=0$ for all $i,n$ in (\ref{lin_observation}). Then, in Fig. \ref{convergence} we report the transient behavior of the squared error $\|\widetilde{\bx}[n]\|^2$ obtained by the ATC algorithm in (\ref{ATC diffusion}), where $\widetilde{\bx}[n]=\{x_i[n]-x_i^o\}_{i=1}^N$, with $x_i[n]$ defined in (\ref{ATC diffusion}) for all $i$. In particular, we report four behaviors, each one associated to a different static sampling strategy (i.e., $p_i=1$ for all $i\in\overline{\S}$), with $|\overline{\S}|$ equal to 3, 5, 10, and 15, respectively. The samples are chosen according to the distributed strategy proposed in Table 2, where the function $f(\cdot)$ is chosen to be the logarithm of the pseudo-determinant. From now on, we will denote this choice as the Max-Det sampling strategy. Also, we set $p_i=1$, and $\sigma_i^2=0$, for all $i$ (because nor noise nor sampling probability play any role in the selection of the samples). An example of graph sampling in the case $|\overline{\S}|=5$ is given in Fig. \ref{fig:Network}, where the black vertices correspond to the sampling nodes. The step-sizes $\mu_i$ in (\ref{ATC diffusion}) are chosen equal to 0.5 for all $i$; the combination weights $\{w_{ij}\}$ are selected using the Metropolis rule \cite{xiao2007distributed}, where $\widetilde{a}_{ij}=1$ if nodes $i$ and $j$ are connected, and $\widetilde{a}_{ij}=0$ otherwise. As we can notice from Fig. \ref{convergence}, as long as condition (\ref{|DcB|<1}) is satisfied (see Sec. III.A), the algorithm drives to zero the error asymptotically, thus perfectly reconstructing the entire signal from a limited number of samples in a totally distributed manner. In particular, as expected, increasing the number of sampling nodes, the learning rate of the algorithm improves. On the contrary, when $|\overline{\S}|<|\F|$, e.g., in the case $|\overline{\S}|=3$, condition (\ref{|DcB|<1}) cannot be satisfied in any way (i.e., the signal is downsampled), and the algorithm cannot reconstruct the graph signal, as shown in Fig. \ref{convergence}.

\begin{figure}[t]
\centering
\includegraphics[width=8.5cm]{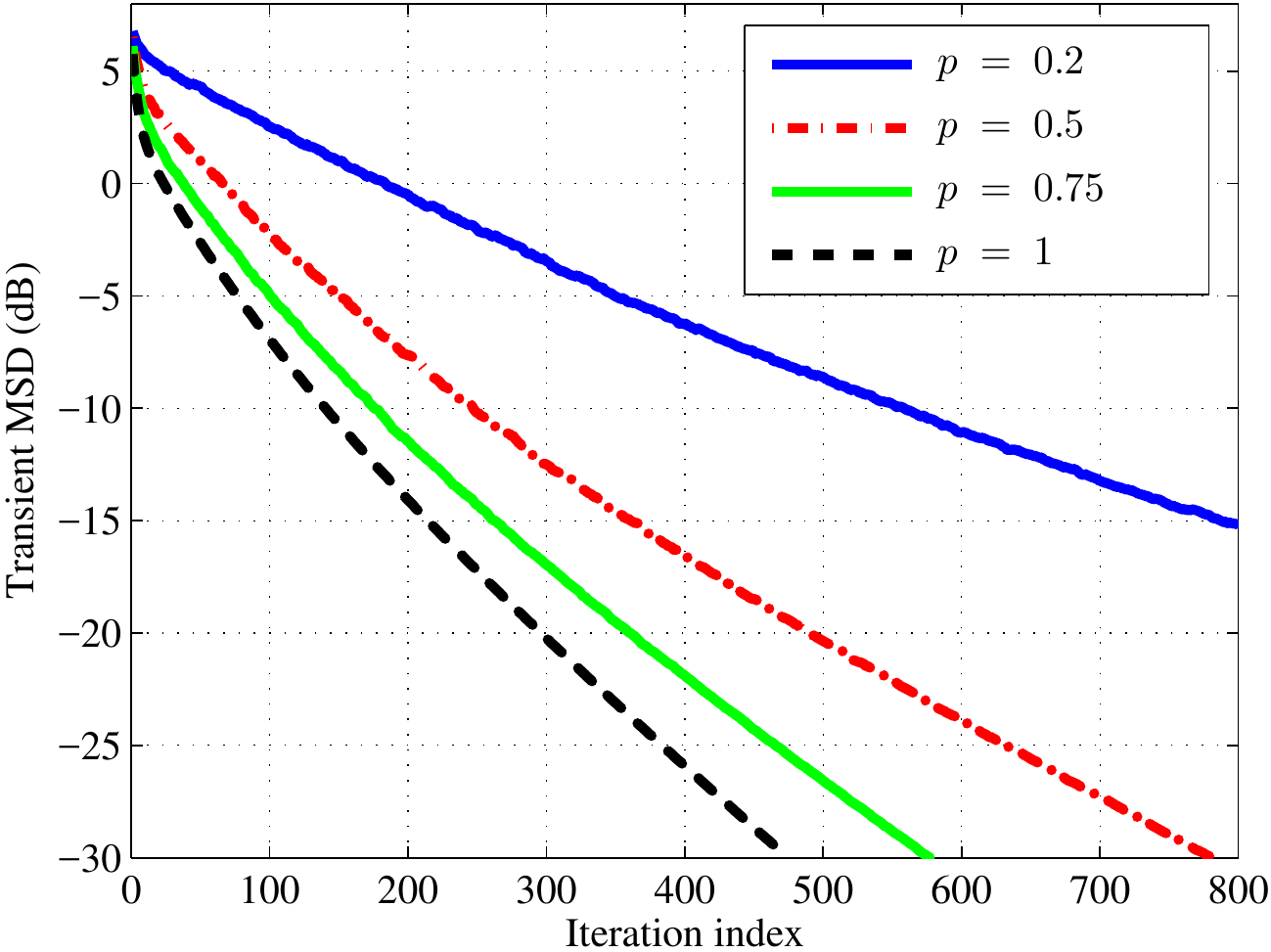}
  \caption{Convergence behavior: Transient MSD in the noise-free case, considering random sampling. $|\mathcal{F}|=5$, $|\overline{\S}|=5$.}\label{convergence2}
\end{figure}
To illustrate the convergence properties of the proposed strategy in the presence of probabilistic sampling (i.e., $0<p_i<1$ for $i\in\overline{\S}$), in Fig. \ref{convergence2} we report the average transient behavior of the squared error $\|\widetilde{\bx}[n]\|^2$ obtained by the ATC algorithm in (\ref{ATC diffusion}), considering different values of sampling probability $p_i=p$ for all $i\in \overline{\mathcal{S}}$. The signal bandwidth is equal to $|\mathcal{F}|=5$, and the expected sampling set is composed of 5 nodes selected according to the Max-Det sampling strategy. The results are averaged over 100 independent simulations.
The step-sizes and the combination weights are chosen as before. As we can notice from Fig. \ref{convergence}, since $\overline{\S}$ satisfies condition (\ref{|DcB|<1}), i.e., the network collects samples from a sufficient number of nodes on average, the algorithm drives to zero the error for any value of $p$. As expected, increasing the sampling probability at each node, the learning rate of the proposed algorithm improves.

\begin{figure}[t]
\centering
\includegraphics[width=8.5cm]{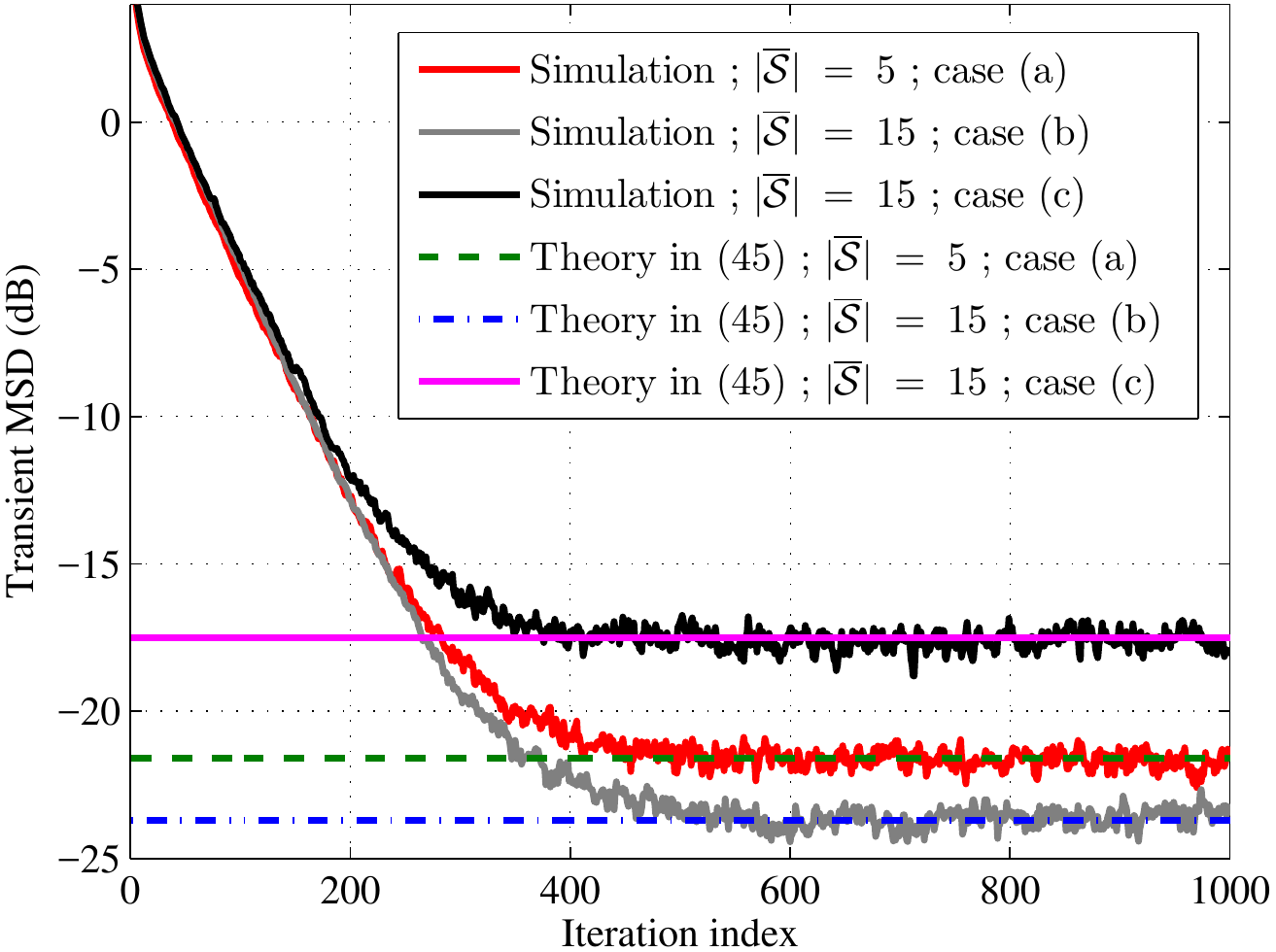}
  \caption{Mean-Square performance: Transient MSD, and theoretical steady-state MSD, for different values of $|\overline{\S}|$. $|\mathcal{F}|=5$.}\label{Trans_MSD}
\end{figure}

\begin{figure}[t]
\centering
\includegraphics[width=8.5cm]{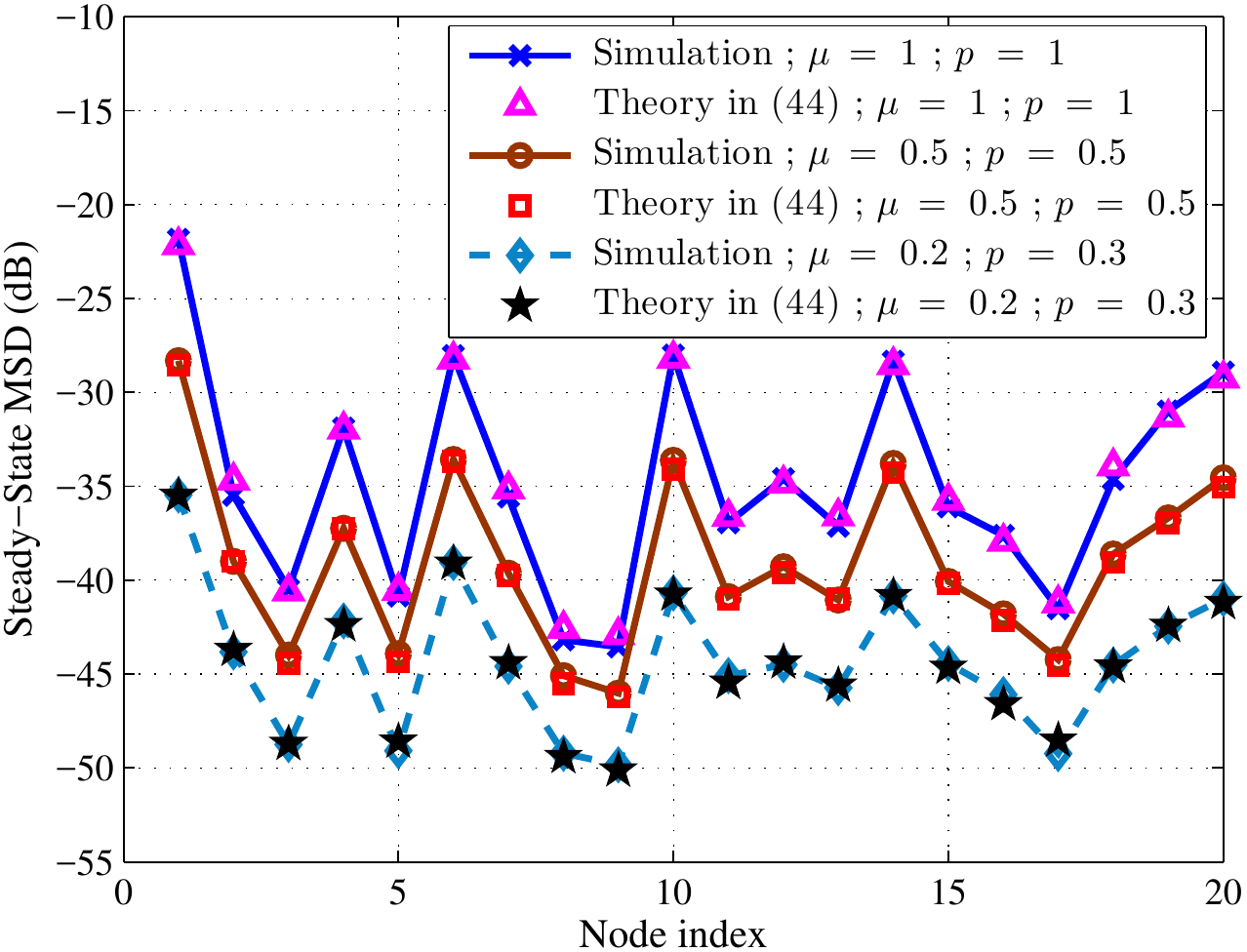}
  \caption{Mean-Square performance: Theoretical and numerical steady-state MSD versus node index. $|\mathcal{F}|=5$, $|\overline{\S}|=10$.}\label{SteadyState_MSD}
\end{figure}

\begin{figure}[t]
\centering
\includegraphics[width=8.5cm]{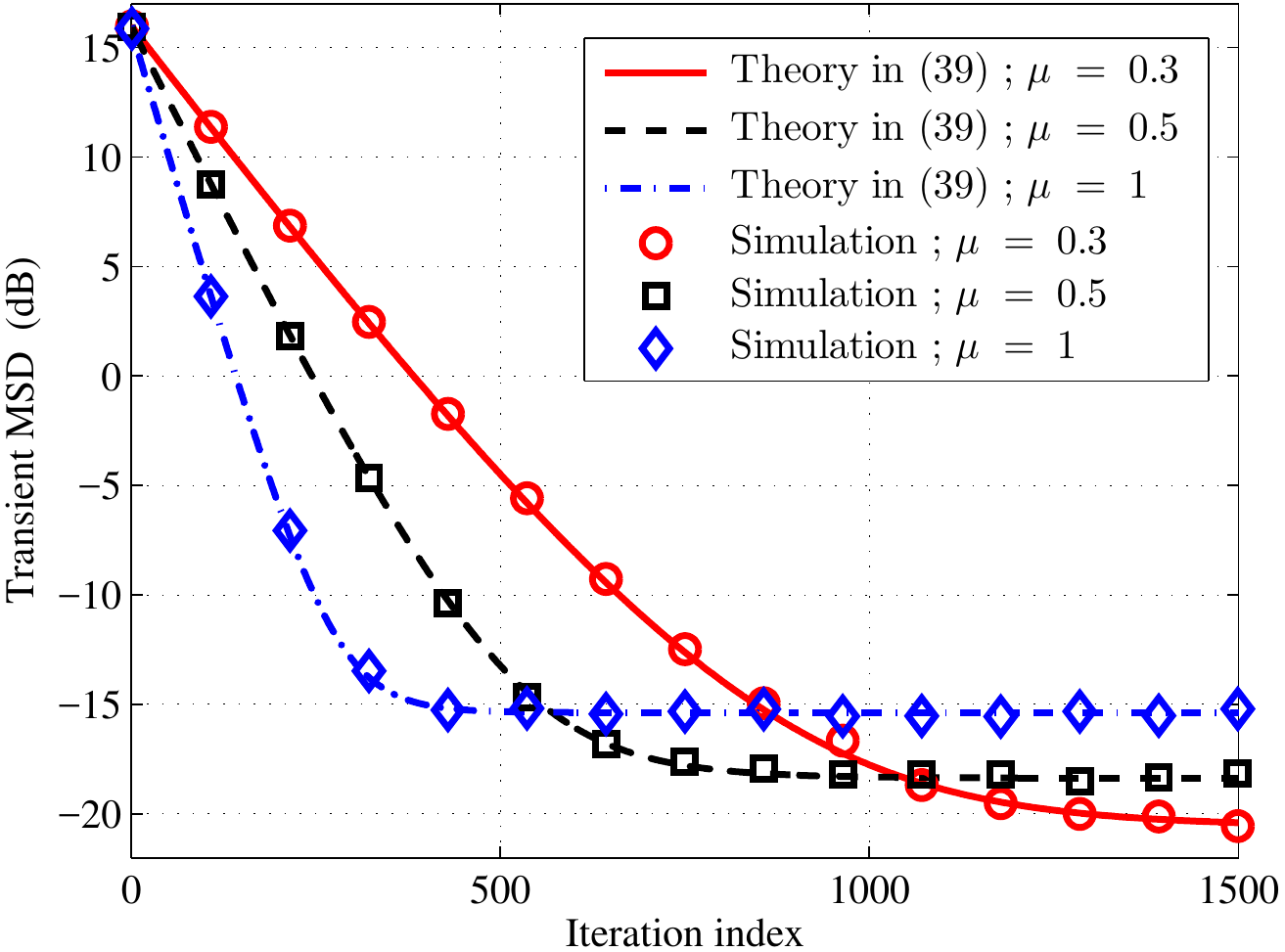}
  \caption{Mean-Square performance: Numerical and theoretical transient MSD, for different values of $\mu$. $|\mathcal{F}|=2$, $|\overline{\S}|=10$.}\label{Trans_MSD_theory}
\end{figure}

\subsubsection{Mean-Square Performance}

Now, we illustrate the mean-square behavior of the proposed strategy in the presence of observation noise in (\ref{lin_observation}). As a first example, we report the transient behavior of the network MSD obtained by the ATC algorithm in (\ref{ATC diffusion}), versus the iteration index, for different number of nodes collecting samples from the network: (a) $|\overline{\mathcal{S}}|=5$; (b) $|\overline{\mathcal{S}}|=15$; (c) $|\overline{\mathcal{S}}|=15$. The difference between the three cases (a), (b) and (c) is also in the observation noise. In particular, in (a) and (b), the noise at the sampling nodes is chosen to be zero-mean, Gaussian, with variance chosen at random between 0 and 0.1. In case (c), the noise variance is chosen equal to case (a) for the first $|\overline{\mathcal{S}}|=5$ nodes belonging also to case (a), whereas it is chosen equal to 0.4 for the remaining 10 sampling nodes. The expected sampling set is chosen according to the Max-Det strategy, and the sampling probabilities are set equal to $p_i=0.8$ for all $i\in\overline{\S}$. The signal bandwidth is equal to $|\mathcal{F}|=5$. The combination weights are chosen as before, and the step-sizes are selected in order to match the learning rates of the algorithm. The curves are averaged over 200 independent simulations, and the corresponding theoretical steady-state values in (\ref{MSD_net}) are reported for the sake of comparison. As we can notice from Fig. \ref{Trans_MSD}, the theoretical predictions match well the simulation results. Furthermore, we notice how, when varying the number of nodes collecting samples, the algorithm might lead to lower or larger steady-state errors. This illustrates that, when reconstructing a graph signal in the presence of noise, it is not always better to increase the number of samples, as this implies an increment of the overall noise injected in the algorithm. In particular, the steady-state performance can improve or degrade by enlarging the sampling set, depending on the distribution of noise over the network. Intuitively, if the noise variance is almost uniform and low at each node of the network, it is convenient to add samples to the algorithm [as from case (a) to case (b)], which improves its learning rate/steady-state performance tradeoff. On the contrary, if some nodes have very noisy observations, it might be not convenient to take their samples (as from case (a) to case (c)), as this might lead to a performance degradation.

As a further example aimed at validating the theoretical results in (\ref{MSDk}), in Fig. \ref{SteadyState_MSD} we report the behavior of the theoretical steady-state MSD values achieved at each vertex of the graph, comparing them with simulation results, for different values of the sampling probability $p$, and of the step-sizes $\mu_i=\mu$ for all $i$. The numerical results are obtained averaging over 200 independent simulations and 500 samples of squared error after convergence of the algorithm.  The signal bandwidth is equal to $|\mathcal{F}|=5$, and the expected sampling set is composed of $|\overline{\S}|=10$ nodes. We can notice from Fig. \ref{SteadyState_MSD} how the theoretical values in (\ref{MSDk}) predict well the simulation results. As expected, reducing the step-size and the sampling probability, the steady-state MSD of the algorithm improves.

Finally, in Fig. \ref{Trans_MSD_theory}, we validate the theoretical expression for the transient MSD in (\ref{weighted_norm3}), comparing it with numerical results, for different values of the step-sizes $\mu_i=\mu$ for all $i$. The numerical results are obtained averaging over 200 independent simulations, the signal bandwidth is equal to $|\mathcal{F}|=2$, $p_i=0.5$ for all $i\in \overline{\mathcal{S}}$, and $|\overline{\S}|=10$ nodes. We can notice from Fig. \ref{Trans_MSD_theory} how the theoretical behaviors in (\ref{weighted_norm3}) predict well the numerical results. As expected, reducing the step-size, the algorithm becomes slower, but the steady-state MSD improves.

\begin{figure}[t]
\centering
\includegraphics[width=8.5cm]{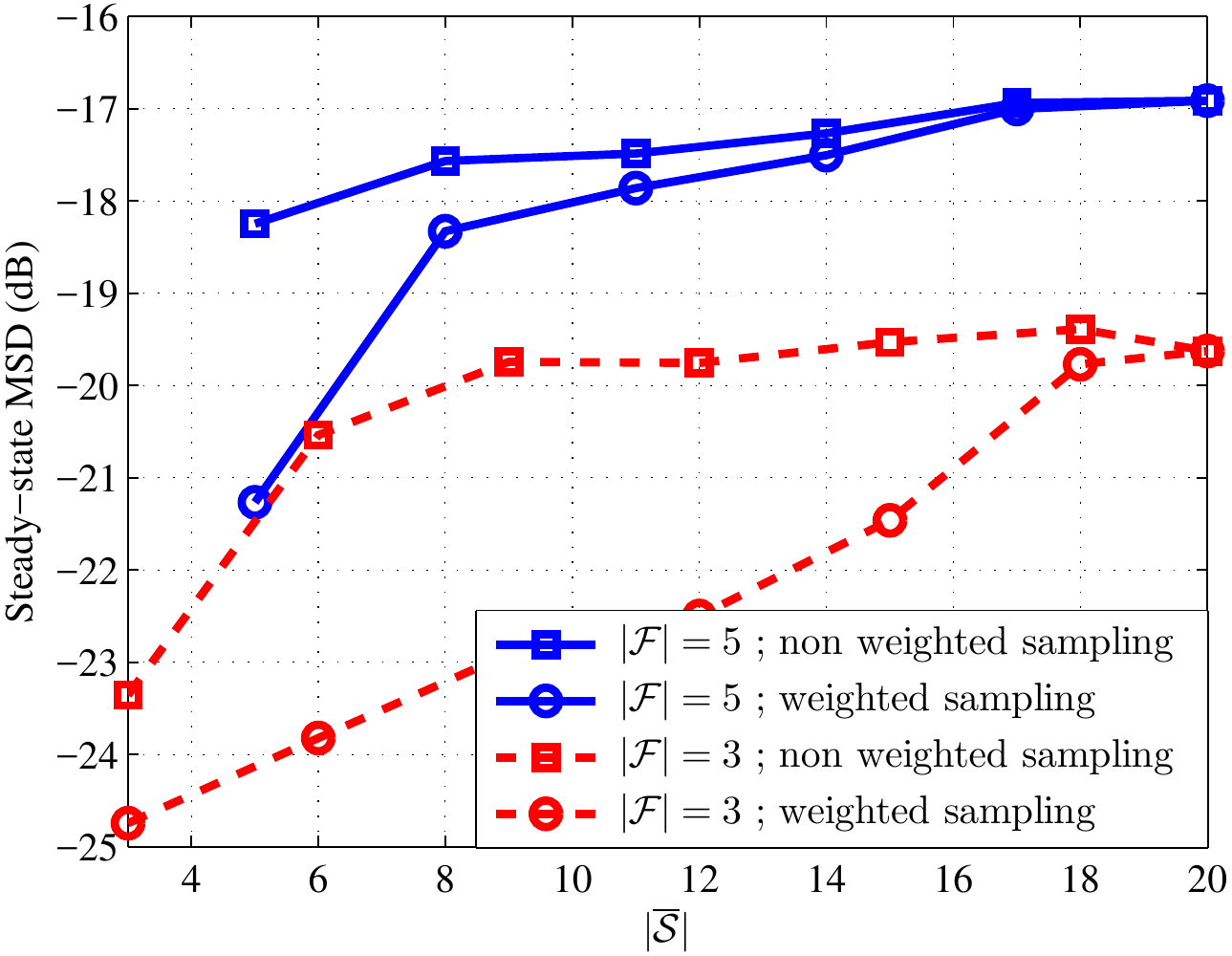}
  \caption{Effect of sampling: Steady-state MSD versus $|\overline{\mathcal{S}}|$, for different graph signal bandwidths and sampling strategies.}\label{sampling}
\end{figure}

\subsubsection{Effect of Sampling Strategy}
As previously remarked, it is fundamental to assess the performance of the algorithm in (\ref{ATC diffusion}) with respect to the strategy adopted to select the expected sampling set $\overline{\mathcal{S}}$. Indeed, when sampling a graph signal, what matters is not only the number of samples, but also (and most important) where the samples are taken. From (\ref{MSD_net}), we can in fact deduce that the sampling set plays a fundamental role, since it affects the performance of the proposed strategy in two ways: (a) it determines the stability of the iteration matrix $\mB$ in (\ref{matrixB}), i.e., $\mH$ in (\ref{MSD_net}); (b) it allows us to select the nodes that inject noise into the system. As a first example, we aim at illustrating the performance obtained by the algorithm in (\ref{ATC diffusion}) under different noise conditions at each node in the network, thus illustrating how selecting samples in a right manner can help reduce the effect of noisy nodes. In particular, we adopt the Max-Det sampling strategy, where the sampling probabilities are set equal to $p_i=0.8$ for all $i\in\overline{\S}$. The noise at each node is chosen to be zero-mean, Gaussian, with a variance chosen uniformly random between 0 and 0.1. The step-sizes are $\mu_i=0.5$ for all $i$, and the combination weights are chosen as before; we also consider the graph in Fig. \ref{fig:Network}. Then, in Fig. \ref{sampling}, we report the steady-state MSD obtained by the algorithm in (\ref{ATC diffusion}), versus $|\overline{\S}|$, for different values of bandwidth $|\F|$ of the graph signal. The curves are averaged over 500 independent simulations. In particular, we consider two variants of the sampling strategy: (a) a weighted strategy as in Table 2, where each local element is weighted by the variance $\sigma_i^2$ of the noise for all $i$ (see, e.g., (\ref{sampling_problem})); and (b) a non-weighted strategy, corresponding to setting $\sigma_i^2=0$ for all $i$, in Table 2. As we can notice from Fig. \ref{sampling}, the weighted strategy always outperforms the non-weighted method; this happens because the weighted strategy tends to select sampling nodes with smaller noise variance, thus leading to better performance. Interestingly, the gain is larger at lower bandwidths, thanks to the larger freedom that the method has in the selection of the (noisy) samples.

\begin{figure}[t]
\centering
\includegraphics[width=8.5cm]{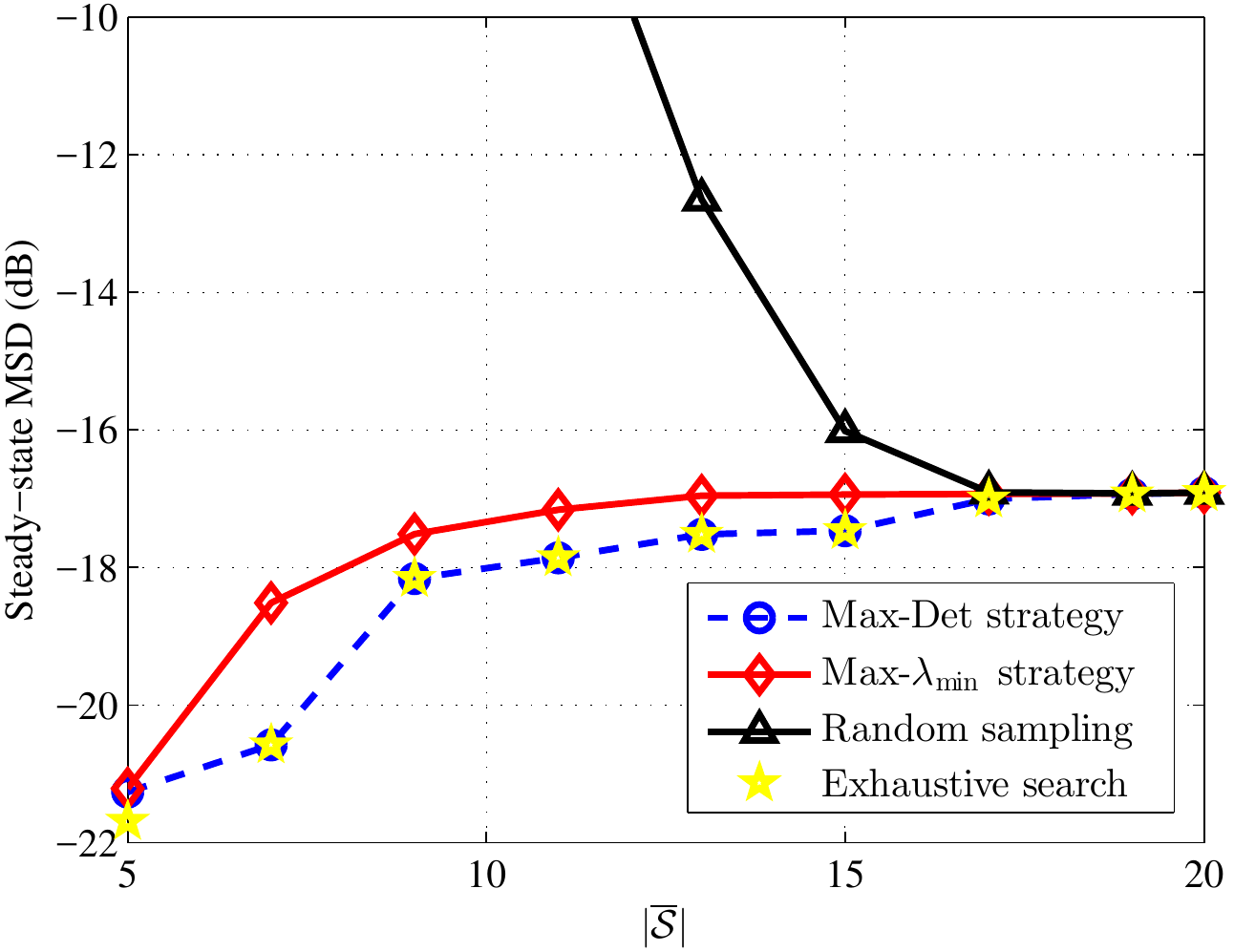}
  \caption{Effect of sampling: Steady-state MSD versus $|\overline{\mathcal{S}}|$, for different sampling strategies.}\label{sampling2}
\end{figure}

\begin{figure}[t]
\centering
\includegraphics[width=8.8cm]{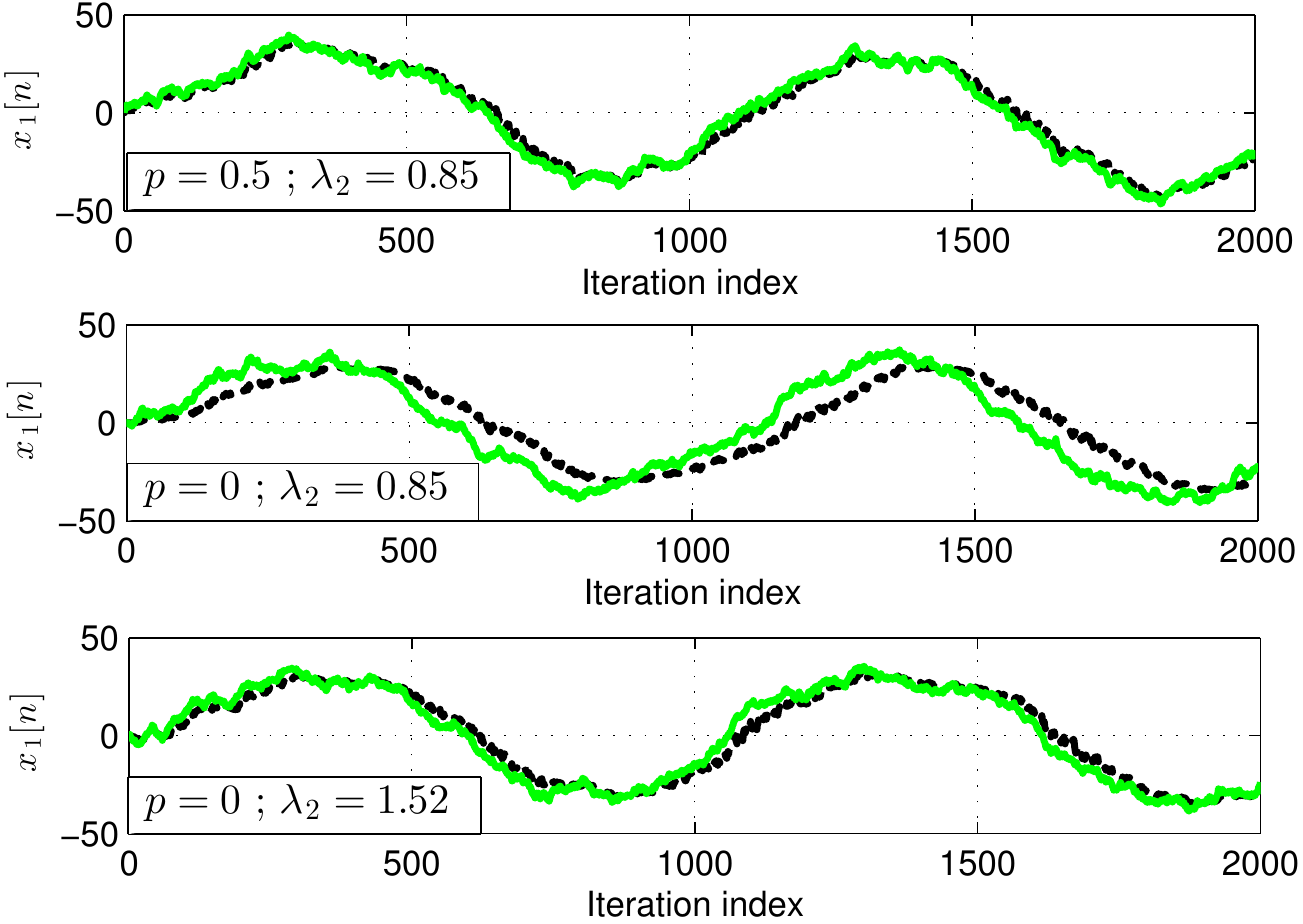}
  \caption{Tracking behavior: Graph signal estimate (dashed) and true signal (solid) versus iteration index, for different values of sampling probability $p$ and graph algebraic connectivity $\lambda_2$.}\label{tracking}
\end{figure}

\begin{figure*}[t]
\centering
\includegraphics[width=18.5cm]{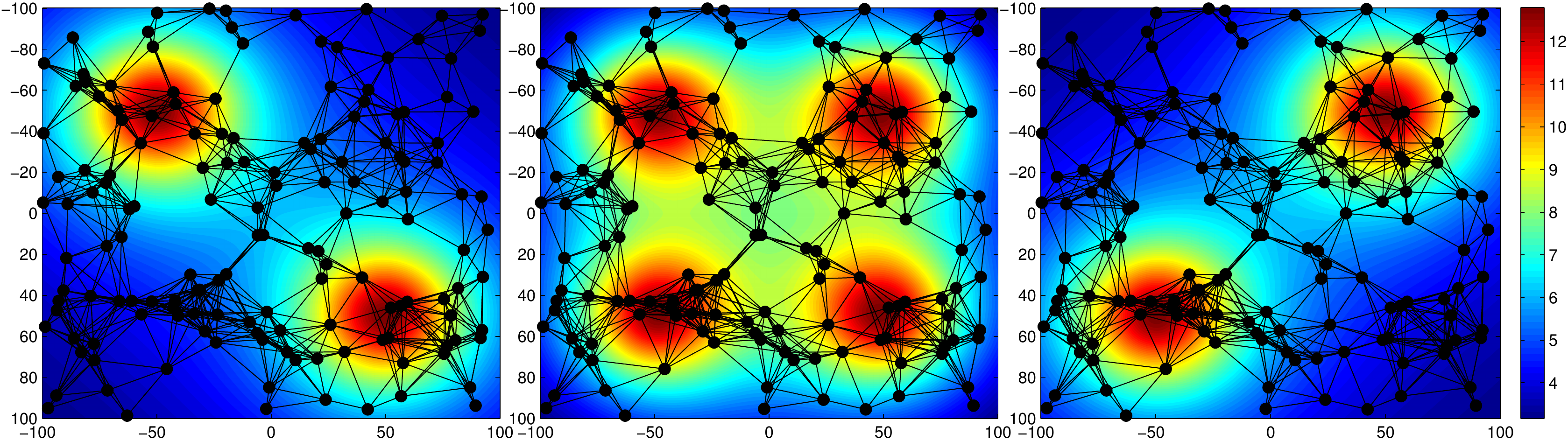}
\caption{PSD estimation in Cognitive Networks: PSD at different time instants, and topology of the cognitive network.}
\label{fig:PF}
\end{figure*}

As a further example, in Fig. \ref{sampling2}, we illustrate the steady-state MSD of the algorithm in (\ref{ATC diffusion}) comparing the performance obtained by four different sampling strategies, namely: (a) the Max-Det strategy (obtained setting $f(\cdot)$ as the logarithm of the pseudo-determinant in Table 2); (b) the Max-$\lambda_{\min}$ strategy (obtained setting $f(\cdot)=\lambda_{\min}(\cdot)$ in Table 2); (c) the random sampling strategy, which simply picks at random $|\overline{\mathcal{S}}|$ nodes; and (d) the exhaustive search procedure aimed at minimizing the MSD in (\ref{MSD_net}) over all the possible sampling combinations. In general, the latter strategy cannot be performed for large graphs and/or in a distributed fashion, and is reported only as a benchmark. We consider a signal bandwidth equal to $|\F|=5$, the sampling probabilities are set equal to $p_i=0.8$ for all $i\in\overline{\S}$, and the results are averaged over 500 independent simulations. The step-sizes and the combination weights are chosen as before. As we can notice from Fig. \ref{sampling2}, the algorithm in (\ref{ATC diffusion}) with random sampling can perform quite poorly, especially at low number of nodes collecting samples. Comparing the other sampling strategies, we notice from Fig. \ref{sampling2} that the Max-Det strategy outperforms all the others, giving good performance also at low number of nodes collecting samples ($|\overline{\mathcal{S}}|=5$ is the minimum number of nodes that allows signal reconstruction). Interestingly, even if the proposed Max-Det strategy is a greedy approach, it shows performance that are comparable to the exhaustive search procedure, which represents the best possible performance achievable by a sampling strategy in terms of MSD. As previously mentioned, this good behavior is due to the monotonicity and sub-modularity properties of the objective function used in the Max-Det strategy, which ensures that the greedy selection strategy in Table 2 achieves performance that are very close to the optimal combinatorial solution \cite{chepuri2015sparsity,shamaiah2010greedy}. Finally, comparing the Max-$\lambda_{\min}$ strategy with the Max-Det strategy, we notice how the latter leads to better performance, because it considers all the modes of the matrix in (\ref{sampling_problem}), as opposed to the single mode associated to the minimum eigenvalue considered by the Max-$\lambda_{\min}$ strategy. This analysis suggests that an optimal design of the sampling strategy for graph signals should take into account processing complexity (in terms of number of nodes collecting samples), prior knowledge (e.g., graph structure, noise distribution), and achievable mean-square performance.

\subsubsection{Tracking of Time-varying Graph Signals}

In this example, we illustrate the tracking capabilities of the proposed distributed methods in the presence of (slowly) time-varying signals evolving over the graph. To this aim, we generate a time-varying signal such that its graph Fourier transform (with respect to the graph in Fig. \ref{fig:Network}, having algebraic connectivity $\lambda_2=0.85$) evolves over time as:
%\begin{align}\label{time_var_sig}
$\bs^o[n+1]=\vartheta\, \bs^o[n]+\bu[n],$
%\end{align}
where $\bs^o[n]\in \mathbb{R}^{|\F|}$, $|\F|=5$, $\vartheta=0.99$, $\bu[n]=\sin(2\pi f_o n)\mathbf{1}+\bw[n]$, $f_o=10^{-3}$, and $\bw[n]$ is a zero-mean, Gaussian noise vector with identity covariance matrix. The corresponding graph signal at time $n$ is then obtained as $\bx^o[n]=\mU_\F\bs^o[n]$. Thus, in Fig. \ref{tracking} (top), we report the behavior of the estimate of the graph signal $x_i[n]$ in (\ref{ATC diffusion}), for $i=1$, using a dashed line. We also report the behavior of the true signal $x^o_i[n]$, using a solid line. The expected sampling set is composed of 10 nodes, and is selected according to the Max-Det sampling strategy; the sampling probabilities are set equal to $p_i=0.5$ for all $i\in\overline{\S}$. In Fig. \ref{tracking} (middle) we repeat the same experiment but setting the sampling probability of node 1 equal to $p_i=0$, i.e., the node never observes the signal. Finally, in Fig. \ref{tracking} (bottom), we consider the case in which the sampling probability of node 1 is equal to zero, but the connectivity of the communication graph linking the nodes is larger than before, having now an algebraic connectivity $\lambda_2=1.52$. The step-sizes are chosen equal to $\mu_i=1$ for all $i$; the combination weights are selected as before. As we can notice from Fig. \ref{tracking}, the algorithm shows good tracking performance in all cases. As expected, the tracking capability is good in the case of Fig. \ref{tracking} (top), when node 1 belongs to the expected sampling set and observes the signal for half of the time. Remarkably, also in the cases of Fig. \ref{tracking} (middle) and (bottom), even if node 1 does not directly observe the signal at its location (i.e., $p=0$), the algorithm can still guarantee good tracking performance thanks to the real-time diffusion of information among nodes in the graph. Finally, comparing Fig. \ref{tracking} (middle) and (bottom), we can notice how a larger connectivity of the communication graph boosts the tracking capabilities of the network thanks to the faster information sharing among the nodes.

%\vspace{-.3cm}
\subsubsection{Application Example - Power Spatial Density Estimation in Cognitive Networks}
In this example, we apply the proposed distributed framework to power density cartography in cognitive radio (CR) networks. We consider a 5G scenario, where a dense deployment of radio access points (RAPs) is envisioned to provide a service environment characterized by very low latency and high rate access. Each RAP collects data related to the transmissions of primary users (PUs) at its geographical position, and communicates with other RAPs with the aim of implementing advanced cooperative sensing techniques. The aim of the CR network is then to build a map of power spatial density (PSD) transmitted by PUs, while processing the received data on the fly and envisaging proper sampling techniques that enable a proactive sensing of the environment from only a limited number of RAP's measurements. \\
\begin{figure}[t]
\centering
\includegraphics[width=8.8cm]{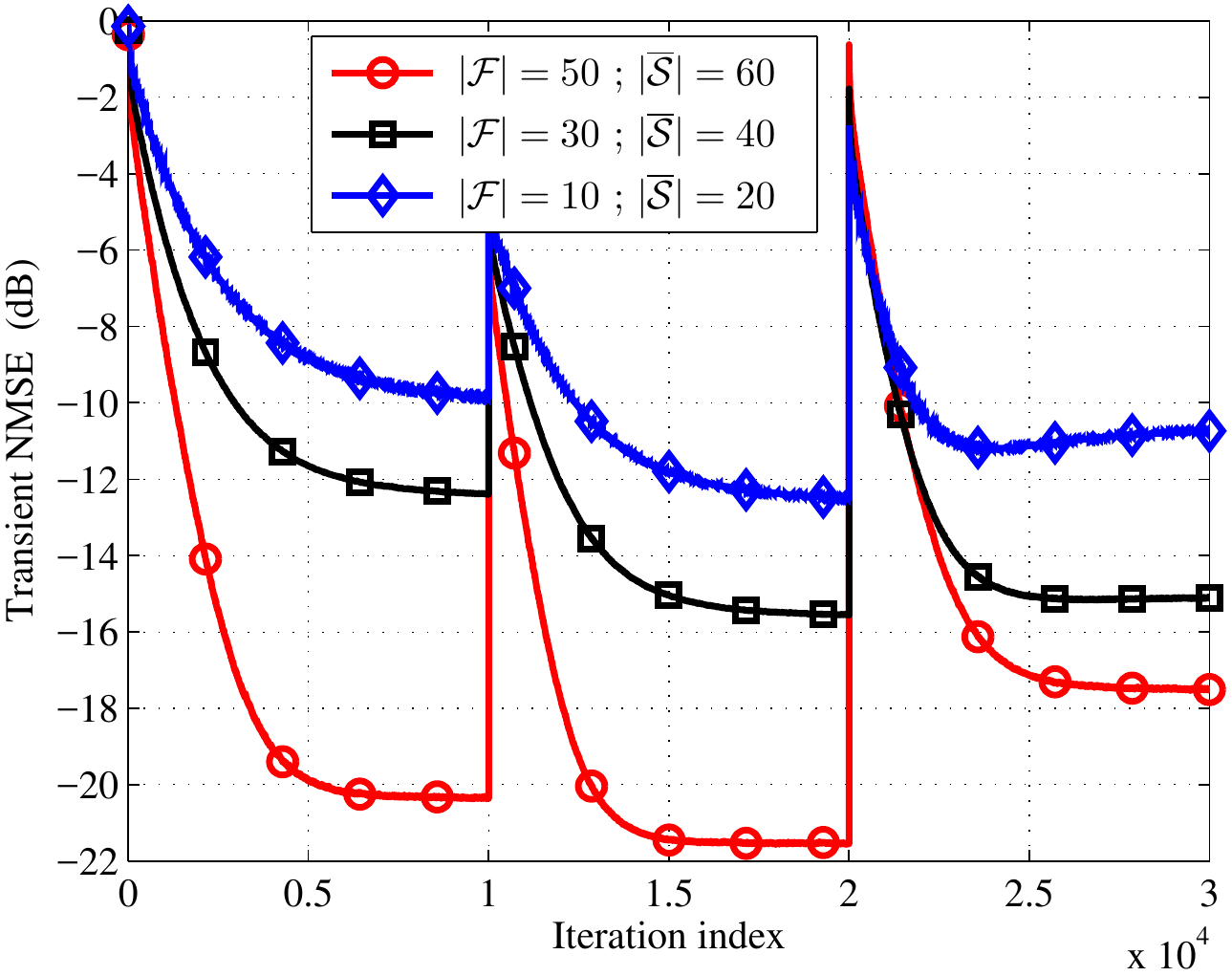}
  \caption{PSD estimation in Cognitive Networks: Transient normalized MSD for different values of $|{\cal F}|$ and $|\overline{{\cal S}}|$.}\label{fig:PF_tracking}
\end{figure}
\indent Let us then consider an operating region of 200 $m^2$ where 150 RAPs are randomly deployed to produce a map of the spatial distribution of power generated by the transmissions of four active primary users. The PU's emit electromagnetic radiation with power equal to 10 mW. The propagation medium is supposed to introduce a free-space path loss attenuation on the PU's transmissions. The graph among RAPs is built from a distance based model, i.e., stations that are sufficiently close to each other are connected through a link. In Fig. \ref{fig:PF}, we illustrate a pictorial description of the scenario, and of the resulting graph signal. For simplicity, we use the graph illustrated in Fig. \ref{fig:PF} for both communication and processing tasks. We assume that each RAP is equipped with an energy detector, which estimates the received signal using 100 samples, considering an additive white Gaussian noise with variance $\sigma_v^2= 10^{-4}$. The resulting signal is not perfectly bandlimited, but it turns out to be smooth over the graph, i.e., neighbor nodes observe similar values. This implies that sampling such signals inevitably introduces aliasing during the reconstruction process. However, even if we cannot find a limited (lower than $N$) set of frequencies where the signal is completely localized, the greatest part of the signal energy is concentrated at low frequencies. This means that if we process the data using a sufficient number of observations and (low) frequencies, we should still be able to reconstruct the signal with a satisfactory performance.\\ \indent An example of PSD cartography based on the proposed diffusion algorithm is shown in Fig. \ref{fig:PF_tracking}, where we simulate a dynamic situation where the four PU's switch between idle and active modes in the order shown in Fig.  \ref{fig:PF} every $10^4$ time instants. In particular, in Fig. \ref{fig:PF_tracking}, we show the behavior of the transient normalized MSD, for different values of $|\overline{\cal S}|$ and bandwidths used for processing. The step-size is chosen equal to 1, the sampling probabilities are $p_i=0.5$ for all $i$, while the adopted sampling strategy is the Max-Det strategy proposed in Table 2. From Fig. \ref{fig:PF_tracking}, we can see how the proposed technique can track time-varying scenarios. Furthermore, as expected, its steady-state performance and learning rate improve with increase in the number of nodes collecting samples and bandwidths used for processing.
%These results illustrate an existing tradeoff between complexity, i.e., number of samples used for processing, and mean-square performance of the proposed strategy. In particular, using a larger bandwidth and a (consequent) larger number of samples for processing, the performance of the algorithm improves, at the price of a larger computational complexity.

\section{Conclusions}

In this paper, we have proposed distributed strategies for adaptive learning of graph signals. The method hinges on the structure of the underlying graph to process data and, under a bandlimited assumption, enables adaptive reconstruction and tracking from a limited number of observations taken over a subset of vertices in a totally distributed fashion. An interesting feature of our proposed method is that the sampling set is allowed to vary over time, and the convergence properties depend only on the expected set of sampling nodes.
Furthermore, the graph topology plays an important role both in the processing and communication aspect of the problem. A detailed mean square analysis is also provided, thus illustrating the role of the sampling strategy on the reconstruction capability, stability, and mean-square performance of the proposed algorithm. Based on this analysis, some useful strategies for the distributed selection of the (expected) sampling set are also provided. Finally, several numerical results are reported to validate the theoretical findings, and illustrate the performance of the proposed method for distributed adaptive learning of signals defined over graphs.

This paper represents the first work that merges the well established field of adaptation and learning over networks, and the emerging topic of signal processing over graphs. Several interesting problems are still open, e.g., distributed reconstruction in the presence of directed and/or switching graph topologies, online identification of the graph signal support from streaming data, distributed inference of the (possibly unknown) graph signal topology, adaptation of the sampling strategy to time-varying scenarios, optimization of the sampling probabilities, just to name a few. We plan to investigate on these exciting problems in our future works.

\section*{Appendix\\ Stability of matrix $\mB$ in (\ref{matrixB})}
Taking the expectation of both sides of (\ref{compact_Diffusion}), and exploiting Assumption 3, we conclude that the mean-error vector evolves according to the following dynamics:
\begin{align}\label{expected_compact_Diffusion}
\mathbb{E}\bee[n+1]=\widehat{\mW}\left(\mI-\mM\widehat{\mP}\mQ\right) \mathbb{E}\bee[n]=\mB\,\mathbb{E}\bee[n].
\end{align}
To prove stability of matrix $\mB$ in (\ref{matrixB}) (and, consequently, the mean stability of the algorithm in (\ref{ATC diffusion})), we proceed by showing that the sequence $\bee[n]$ in (\ref{expected_compact_Diffusion}) asymptotically vanishes for any initial condition. To this aim, let $\by[n]=\mathbb{E} \bee[n]$, and consider its decomposition as:
\begin{eqnarray}
  \by[n] = \overline{\by}[n]+\widetilde{\by}[n], \label{y_dec}
\end{eqnarray}
where $\overline{\by}[n]$ represents the average vector over all nodes, and $\widetilde{\by}[n]$ is a disagreement error, respectively given by:
\begin{eqnarray}
  &&\overline{\by}[n] =\mJ\, \by[n]= (\mathbf{1}\otimes \mI)\,\hat{\by}[n],   \label{y_bar}\\
  && \widetilde{\by}[n]=\mJ_\perp\,\by[n], \label{y_tilde}
\end{eqnarray}
with
\begin{align}
&\hat{\by}[n]=\frac{1}{N}\sum_{i=1}^N\by_i[n],\label{y_hat}\\
&\mJ=\frac{1}{N}\,\mathbf{1}\mathbf{1}^T\otimes \mI, \quad \hbox{and} \quad \mJ_\perp=\mI-\mJ. \label{J}
\end{align}
In the sequel, we will show that both $\overline{\by}[n]$ (or, equivalently, $\hat{\by}[n]$) and $\widetilde{\by}[n]$ asymptotically converge to zero, thus proving the convergence in the mean of the algorithm and the stability of matrix $\mB$ in (\ref{matrixB}). From (\ref{y_bar}) and (\ref{expected_compact_Diffusion}), we obtain
\begin{align}\label{y_bar_recursion}
\overline{\by}[n+1] \;&=\; \mJ\widehat{\mW}\left(\mI-\mM\widehat{\mP}\mQ\right) \by[n]\nonumber\\
\;& \stackrel{(a)}{=}\; \mJ\by[n]-\mJ\mM\widehat{\mP}\mQ\by[n]\nonumber\\
%\;& \stackrel{(a)}{=}\; \frac{1}{N}\left(\mathbf{1}^T\otimes \mI\right)\left(\mI-\mM\widehat{\mP}\mQ\right) \by[n] \nonumber\\
%\;& \stackrel{(b)}{=}\; \overline{\by}[n]-\frac{1}{N}\left(\mathbf{1}^T\otimes \mI\right)\mM\widehat{\mP}\mQ \,\by[n] \nonumber\\
 \;& \stackrel{(b)}{=}\; \left(\mI-\mJ\mM\widehat{\mP}\mQ\right)\overline{\by}[n]-\mJ\mM\widehat{\mP}\mQ \,\widetilde{\by}[n]
%\;&\qquad-\frac{1}{N}\left(\mathbf{1}^T\otimes \mI\right)\mM\widehat{\mP}\mQ \,\widetilde{\by}[n]
\end{align}
where in (a) we have used $\mJ\widehat{\mW}=\mJ$ [cf. (\ref{combination_coefficients}), (\ref{combination_matrices}) and (\ref{J})]; and in (b) we have exploited (\ref{y_bar}) and (\ref{y_dec}). Similarly, from (\ref{y_tilde}) and (\ref{expected_compact_Diffusion}), we get
\begin{align}\label{y_tilde_recursion}
\widetilde{\by}[n+1] &\;=\; \mJ_\perp\widehat{\mW}\left(\mI-\mM\widehat{\mP}\mQ\right) \by[n]\nonumber\\
 &\hspace{-.6cm}\stackrel{(a)}{=}\;  \mJ_\perp\widehat{\mW}\mJ_\perp\by[n]-\mJ_\perp\widehat{\mW}\mM\widehat{\mP}\mQ\,\by[n] \nonumber\\
 &\hspace{-.6cm}\stackrel{(b)}{=} \mJ_\perp\widehat{\mW}\left(\mI-\mM\widehat{\mP}\mQ\right)\widetilde{\by}[n]-\mJ_\perp\widehat{\mW}\mM\widehat{\mP}\mQ\, \overline{\by}[n]
\end{align}
where in (a) we have exploited the relation $\mJ_\perp\widehat{\mW}=\mJ_\perp\widehat{\mW}\mJ_\perp$ [cf. (\ref{combination_coefficients}), (\ref{combination_matrices}) and (\ref{J})]; and in (b) we have used (\ref{y_tilde}) and (\ref{y_dec}). Now, combining the recursions (\ref{y_bar_recursion}) and (\ref{y_tilde_recursion}), we obtain
\begin{align}\label{error_recursion}
\begin{bmatrix} \overline{\by}[n+1] \vspace{.2cm}\\ \widetilde{\by}[n+1] \end{bmatrix}=\begin{pmatrix}
                                                                            \mZ_{11} & \mZ_{12}  \vspace{.2cm}\\
                                                                             \mZ_{21} &  \mZ_{22}\\
                                                                          \end{pmatrix}
\begin{bmatrix} \overline{\by}[n] \vspace{.2cm}\\ \widetilde{\by}[n] \end{bmatrix},
\end{align}
where
\begin{align}
%& \mZ_{11} \,=\, \mI-\frac{1}{N}\left(\mathbf{1}^T\otimes \mI\right)\mM\widehat{\mP}\mQ\left(\mathbf{1}\otimes \mI\right),\label{Z11}\\
&\mZ_{11} \,=\, \mI-\mJ\mM\widehat{\mP}\mQ,\label{Z11}\\
%& \mZ_{12} \,=\, -\frac{1}{N}\left(\mathbf{1}^T\otimes \mI\right)\mM\widehat{\mP}\mQ,\label{Z12}\\
& \mZ_{12} \,=\, -\mJ\mM\widehat{\mP}\mQ,\label{Z12}\\
%& \mZ_{21} \,=\, -\mJ_\perp\widehat{\mW}\mM\widehat{\mP}\mQ\left(\mathbf{1}\otimes \mI\right),\label{Z21}\\
& \mZ_{21} \,=\, -\mJ_\perp\widehat{\mW}\mM\widehat{\mP}\mQ,\label{Z21}\\
& \mZ_{22} \,=\, \mJ_\perp\widehat{\mW}\left(\mI-\mM\widehat{\mP}\mQ\right)\label{Z22}.
\end{align}
A necessary and sufficient condition that guarantee the convergence to zero of the sequence in (\ref{error_recursion}) is that matrix
\begin{align}\label{matrix_Z}
\mZ=\begin{pmatrix}
   \mZ_{11} & \mZ_{12}  \vspace{.2cm}\\
   \mZ_{21} &  \mZ_{22}\\
\end{pmatrix}
\end{align}
is stable \cite{horn2012matrix}. We proceed by showing that, under Assumption 4, the eigenvalues of matrix $\mZ$ in (\ref{matrix_Z}) are approximatively determined only by the eigenvalues of $\mZ_{11}$ and $\mZ_{22}$. From (\ref{matrix_Z}), the characteristic polynomial of $\mZ$ is given by:
\begin{align}\label{car_polynomial}
p(\lambda)&=\det\left(\mZ-\lambda\mI\right) \nonumber \\
&\stackrel{(a)}{=}\, \det\big((\mZ_{22}-\lambda\mI)(\mZ_{11}-\lambda\mI)-\mZ_{21}\mZ_{12})\big)\nonumber\\
&\stackrel{(b)}{\simeq}\, \det(\mZ_{22}-\lambda\mI)\det(\mZ_{11}-\lambda\mI)
\end{align}
where $(a)$ holds for $2\times2$ block matrices \cite[p.4]{silvester2000determinants}, since $\mZ_{11}-\lambda\mI$ and $\mZ_{12}$ commute [cf. (\ref{Z11}) and (\ref{Z12})];
and $(b)$ follows from the small-step size Assumption 4, as proved next. Indeed, expanding the argument of the determinant in (\ref{car_polynomial}a) we obtain:
\begin{align}\label{p1}
&(\mZ_{22}-\lambda\mI)(\mZ_{11}-\lambda\mI)-\mZ_{21}\mZ_{12}\nonumber\\
&\qquad=\mZ_{22}\mZ_{11}-\mZ_{21}\mZ_{12}-(\mZ_{22}+\mZ_{11})\lambda+\lambda^2\mI.
\end{align}
Thus, if under Assumption 4 we have
\begin{align}\label{p2}
\mZ_{22}\mZ_{11}-\mZ_{21}\mZ_{12}\approx \mZ_{22}\mZ_{11},
\end{align}
from (\ref{p1}) and (\ref{car_polynomial}a), we can conclude that (\ref{car_polynomial}b) holds, i.e., $$(\mZ_{22}-\lambda\mI)(\mZ_{11}-\lambda\mI)-\mZ_{21}\mZ_{12}\approx (\mZ_{22}-\lambda\mI)(\mZ_{11}-\lambda\mI).$$
Now, from (\ref{Z11})-(\ref{Z22}), we easily obtain:
\begin{align}
&\mZ_{22}\mZ_{11}=\mJ_\perp\widehat{\mW}-\mJ_\perp\widehat{\mW}\mJ\mM\widehat{\mP}\mQ-\mJ_\perp\widehat{\mW}\mM\widehat{\mP}\mQ \nonumber\\
&\qquad\qquad+\mJ_\perp\widehat{\mW}\mM\widehat{\mP}\mQ\mJ\mM\widehat{\mP}\mQ\label{p3}\\
&\mZ_{21}\mZ_{12}= \mJ_\perp\widehat{\mW}\mM\widehat{\mP}\mQ\mJ\mM\widehat{\mP}\mQ \nonumber\\
&\mZ_{22}\mZ_{11}-\mZ_{21}\mZ_{12}=\mJ_\perp\widehat{\mW}-\mJ_\perp\widehat{\mW}\mJ\mM\widehat{\mP}\mQ-\mJ_\perp\widehat{\mW}\mM\widehat{\mP}\mQ \nonumber
\end{align}
It is then clear that, using Assumption 4 and thus neglecting the term $\mJ_\perp\widehat{\mW}\mM\widehat{\mP}\mQ\mJ\mM\widehat{\mP}\mQ=O(\mu_{\max}^2)$ in (\ref{p3}) with respect to the constant term and the term $O(\mu_{\max})$ contained in the expression of $\mZ_{22}\mZ_{11}$, we obtain (\ref{p2}). As previously mentioned, this proves that the approximation made in (\ref{car_polynomial}b) holds under the small step-sizes Assumption 4.

From (\ref{car_polynomial}), we conclude that, for sufficiently small step-sizes, the eigenvalues of matrix $\mZ$ in (\ref{matrix_Z}) are approximatively given by the eigenvalues of $\mZ_{11}$ and $\mZ_{22}$ in (\ref{Z11}) and (\ref{Z22}), i.e. matrix $\mZ$ is stable if matrices $\mZ_{11}$  and $\mZ_{22}$ are also stable. This means that the iteration matrix in (\ref{error_recursion}) can be considered approximatively diagonal for the purpose of stability analysis. Thus, in the sequel, we analyze the stability of the recursion in (\ref{error_recursion}), considering separately the behavior of the mean vector $\overline{\by}[n]$ and of the fluctuation $\widetilde{\by}[n]$, under the aforementioned diagonal approximation.

\textit{Convergence of $\overline{\by}[n]$:} We now study the recursion $$\overline{\by}[n+1]=\mZ_{11}\,\overline{\by}[n].$$ For convenience, exploiting (\ref{y_bar}), (\ref{Z11}), and (\ref{J}), we equivalently recast the previous recursion in terms of $\hat{\by}[n]$, as:
\begin{align}\label{y_hat}
\hat{\by}[n+1]=\left(\mI-\frac{1}{N}\left(\mathbf{1}^T\otimes \mI\right)\mM\widehat{\mP}\mQ\left(\mathbf{1}\otimes \mI\right)\right)\hat{\by}[n].
\end{align}
The recursion (\ref{y_hat}) converges to zero if the two following conditions hold: (a) matrix
\begin{align}\label{cond3}
\mV=\frac{1}{N}\left(\mathbf{1}^T\otimes \mI\right)\mM\widehat{\mP}\mQ\left(\mathbf{1}\otimes \mI\right)=\frac{1}{N}\sum_{i\in\overline{\S}} \mu_i p_i \bc_i\bc_i^H
\end{align}
is invertible (i.e., full rank); (b) and $|1-\lambda_{\max}(\mV)|<1$. Proceeding as in (\ref{cond2})-(\ref{|DcB|<1}), the invertibility of matrix (\ref{cond3}) is guaranteed under condition (\ref{|DcB|<1}). Then, if matrix (\ref{cond3}) is full rank, exploiting the inequality
\begin{align}
\lambda_{\max}(\mV)\,=\, \frac{1}{N}\left\|\sum_{i\in\overline{\S}}\, \mu_i p_i\bc_i\bc_i^H\right\|\,\leq\, \frac{\mu_{\max}}{N}\sum_{i\in\overline{\S}}p_i\|\bc_{i}\|^2,\nonumber
\end{align}
condition (b) is guaranteed if the step-sizes satisfy:
\begin{eqnarray}\label{step_sizes2}
0<\mu_i\leq\mu_{\max}<\frac{2}{\displaystyle\frac{1}{N}\sum_{i\in\overline{\S}}p_i\|\bc_{i}\|^2}, \quad \hbox{for all $i\in \overline{\mathcal{S}}$},
\end{eqnarray}
which hold true under Assumption 4. Thus, under conditions (\ref{|DcB|<1}) and assumption 4, $\hat{\by}[n]$ (and $\overline{\by}[n]$) converges to zero for all initial conditions, i.e., matrix $\mZ_{11}$ is stable.

\textit{Convergence of $\widetilde{\by}[n]$:} We now study the recursion $$\widetilde{\by}[n+1]=\mZ_{22}\,\widetilde{\by}[n],$$
which converges to zero if $\mZ_{22}$ is stable. From (\ref{Z22}), we have
\begin{align}\label{cond4}
\rho(\mZ_{22})\leq \left\|\mJ_\perp\widehat{\mW}\right\|\left\|\mI-\mM\widehat{\mP}\mQ\right\|,
\end{align}
with $\rho(\mX)$ denoting the spectral radius of a matrix $\mX$. Under Assumption 2, we have [cf. (\ref{combination_matrices}) and (\ref{J})]
\begin{align}\label{cond5}
\left\|\mJ_\perp\widehat{\mW}\right\|=\left\|\left(\mW-\frac{1}{N}\mathbf{1}\mathbf{1}^T\right)\otimes \mI\right\| <1.
\end{align}
Thus, from (\ref{cond4}) and (\ref{cond5}), $\rho(\mZ_{22})<1$, i.e., matrix $\mZ_{22}$ in (\ref{Z22}) is stable, if $\left\|\mI-\mM\widehat{\mP}\mQ\right\|\leq1$, which holds true under assumption 4.
%\begin{eqnarray}\label{step_sizes3}
%0<\mu_i<\frac{2}{p_i\|\bc_i\|^2}, \quad \hbox{for all $i\in \overline{\mathcal{S}}$},
%\end{eqnarray}
%i.e.,
In conclusion, matrix $\mZ$ in (\ref{matrix_Z}) is stable, and the sequence $\by[n]$ in (\ref{y_dec}) [i.e., $\mathbb{E}\bee[n]$ in (\ref{expected_compact_Diffusion})] asymptotically vanishes for all possible initial conditions. This proves the stability of matrix $\mB$ in (\ref{matrixB}).

\section*{Acknowledgement}

The authors would like thank the anonymous reviewers for the detailed suggestions that improved the manuscript.

%\balance
\bibliographystyle{MyIEEE}
\bibliography{refs}

\begin{thebibliography}{10}
\providecommand{\url}[1]{#1}
\csname url@samestyle\endcsname
\providecommand{\newblock}{\relax}
\providecommand{\bibinfo}[2]{#2}
\providecommand{\BIBentrySTDinterwordspacing}{\spaceskip=0pt\relax}
\providecommand{\BIBentryALTinterwordstretchfactor}{4}
\providecommand{\BIBentryALTinterwordspacing}{\spaceskip=\fontdimen2\font plus
\BIBentryALTinterwordstretchfactor\fontdimen3\font minus
  \fontdimen4\font\relax}
\providecommand{\BIBforeignlanguage}[2]{{%
\expandafter\ifx\csname l@#1\endcsname\relax
\typeout{** WARNING: IEEEtran.bst: No hyphenation pattern has been}%
\typeout{** loaded for the language `#1'. Using the pattern for}%
\typeout{** the default language instead.}%
\else
\language=\csname l@#1\endcsname
\fi
#2}}
\providecommand{\BIBdecl}{\relax}
\BIBdecl

\bibitem{shuman2013emerging}
D.~I. Shuman, S.~K. Narang, P.~Frossard, A.~Ortega, and P.~Vandergheynst, ``The
  emerging field of signal processing on graphs: Extending high-dimensional
  data analysis to networks and other irregular domains,'' \emph{IEEE Signal
  Proc. Mag.}, vol.~30, no.~3, pp. 83--98, 2013.

\bibitem{sandryhaila2013discrete}
A.~Sandryhaila and J.~M.~F. Moura, ``Discrete signal processing on graphs,''
  \emph{IEEE Trans. on Sig. Proc.}, vol.~61, no.~7, pp. 1644--1656, 2013.

\bibitem{sandryhaila2014big}
------, ``Big data analysis with signal processing on graphs: Representation
  and processing of massive data sets with irregular structure,'' \emph{IEEE
  Signal Proc. Mag.}, vol.~31, no.~5, pp. 80--90, 2014.

\bibitem{sandryhaila2014discrete}
A.~Sandryhaila and J.~M. Moura, ``Discrete signal processing on graphs:
  Frequency analysis,'' \emph{IEEE Transactions on Signal Processing}, vol.~62,
  no.~12, pp. 3042--3054, 2014.

\bibitem{narang2012perfect}
S.~K. Narang and A.~Ortega, ``Perfect reconstruction two-channel wavelet filter
  banks for graph structured data,'' \emph{IEEE Transactions on Signal
  Processing}, vol.~60, no.~6, pp. 2786--2799, 2012.

\bibitem{narang2013compact}
------, ``Compact support biorthogonal wavelet filterbanks for arbitrary
  undirected graphs,'' \emph{IEEE Transactions on Signal Processing}, vol.~61,
  no.~19, pp. 4673--4685, 2013.

\bibitem{pesenson2008sampling}
I.~Z. Pesenson, ``Sampling in {Paley-Wiener} spaces on combinatorial graphs,''
  \emph{Trans. of the American Mathematical Society}, vol. 360, no.~10, pp.
  5603--5627, 2008.

\bibitem{zhu2012approximating}
X.~Zhu and M.~Rabbat, ``Approximating signals supported on graphs,'' in
  \emph{IEEE Int. Conf. on Acoustics, Speech and Signal Processing (ICASSP)},
  Kyoto, March 2012, pp. 3921--3924.

\bibitem{chen2015discrete}
S.~Chen, R.~Varma, A.~Sandryhaila, and J.~Kova{\v c}evi{\'c}, ``Discrete signal
  processing on graphs: Sampling theory,'' \emph{IEEE Trans. on Signal Proc.},
  vol.~63, no.~24, pp. 6510--6523, Dec. 2015.

\bibitem{Puschel1}
M.~P{\"u}schel and J.~M.~F. Moura, ``Algebraic signal processing theory:
  Foundation and 1-{D} time,'' \emph{{IEEE Trans. Signal Process.}}, vol.~56,
  no.~8, pp. 3572--3585, 2008.

\bibitem{Puschel2}
------, ``Algebraic signal processing theory: 1-{D} space,'' \emph{IEEE Trans.
  on Signal Processing}, vol.~56, no.~8, pp. 3586--3599, 2008.

\bibitem{narang2013signal}
S.~Narang, A.~Gadde, and A.~Ortega, ``Signal processing techniques for
  interpolation in graph structured data,'' in \emph{IEEE International
  Conference on Acoustics, Speech and Signal Processing (ICASSP)}, Vancouver,
  May 2013, pp. 5445--5449.

\bibitem{tsitsvero2015signals}
M.~Tsitsvero, S.~Barbarossa, and P.~Di~Lorenzo, ``Signals on graphs:
  Uncertainty principle and sampling,'' \emph{IEEE Transactions on Signal
  Processing}, vol.~64, no.~18, pp. 4845--4860, 2016.

\bibitem{wang2014local}
X.~Wang, P.~Liu, and Y.~Gu, ``Local-set-based graph signal reconstruction,''
  \emph{IEEE Trans. on Signal Proc.}, vol.~63, no.~9, pp. 2432--2444, 2015.

\bibitem{marquez2015}
A.~G. Marquez, S.~Segarra, G.~Leus, and A.~Ribeiro, ``Sampling of graph signals
  with successive local aggregations,'' \emph{IEEE Transaction on Signal
  Processing}, vol.~65, no.~7, pp. 1832--1843, Apr. 2016.

\bibitem{TsitsveroEusipco15}
M.~Tsitsvero and S.~Barbarossa, ``On the degrees of freedom of signals on
  graphs,'' in \emph{European Signal Processing Conference}, Nice, Sept 2015,
  pp. 1521--1525.

\bibitem{narang2013localized}
S.~K. Narang, A.~Gadde, E.~Sanou, and A.~Ortega, ``Localized iterative methods
  for interpolation in graph structured data,'' in \emph{IEEE Global Conference
  on Signal and Information Processing}, Austin, Dec. 2013, pp. 491--494.

\bibitem{segarra2015reconstruction}
S.~Segarra, A.~G. Marques, G.~Leus, and A.~Ribeiro, ``Reconstruction of graph
  signals through percolation from seeding nodes,'' \emph{IEEE Trans. on Signal
  Proc.}, vol.~64, no.~16, pp. 4363--4378, Aug. 2016.

\bibitem{sandryhaila2013classification}
A.~Sandryhaila and J.~M. Moura, ``Classification via regularization on
  graphs.'' in \emph{Proc. of IEEE Global conference on Signal and Information
  Processing}, Austin, Dec. 2013, pp. 495--498.

\bibitem{thanou2013parametric}
D.~Thanou, D.~I. Shuman, and P.~Frossard, ``Parametric dictionary learning for
  graph signals,'' in \emph{IEEE Global Conference on Signal and Information
  Processing}, Austin, Dec. 2013, pp. 487--490.

\bibitem{zhou2004regularization}
D.~Zhou and B.~Sch{\"o}lkopf, ``A regularization framework for learning from
  graph data,'' in \emph{ICML workshop on statistical relational learning and
  Its connections to other fields}, vol.~15, July 2004, pp. 67--68.

\bibitem{belkin2006manifold}
M.~Belkin, P.~Niyogi, and V.~Sindhwani, ``Manifold regularization: A geometric
  framework for learning from labeled and unlabeled examples,'' \emph{The J. of
  Machine Learning Research}, vol.~7, pp. 2399--2434, 2006.

\bibitem{chen2015signal}
S.~Chen, A.~Sandryhaila, J.~M. Moura, and J.~Kovacevic, ``Signal recovery on
  graphs: Variation minimization,'' \emph{IEEE Transactions on Signal
  Processing}, vol.~63, no.~17, pp. 4609--4624, 2015.

\bibitem{chen2015signalrecovery}
S.~Chen, R.~Varma, A.~Singh, and J.~Kova{\v{c}}evi{\'c}, ``Signal recovery on
  graphs: Fundamental limits of sampling strategies,'' \emph{IEEE Trans. on
  Signal and Inf. Proc. over Networks}, vol.~2, no.~4, pp. 539--554, Dec. 2016.

\bibitem{chen2014signal}
S.~Chen, A.~Sandryhaila, G.~Lederman, Z.~Wang, J.~M. Moura, P.~Rizzo,
  J.~Bielak, J.~H. Garrett, and J.~Kovacevic, ``Signal inpainting on graphs via
  total variation minimization,'' in \emph{IEEE Conference on Acoustics, Speech
  and Signal Processing}, Florence, May 2014, pp. 8267--8271.

\bibitem{chen2014signaldenoising}
S.~Chen, A.~Sandryhaila, J.~M. Moura, and J.~Kovacevic, ``Signal denoising on
  graphs via graph filtering,'' in \emph{IEEE Global Conference on Signal and
  Information Processing}, Atlanta, Dec. 2014, pp. 872--876.

\bibitem{di2016least}
P.~Di~Lorenzo, S.~Barbarossa, P.~Banelli, and S.~Sardellitti, ``Adaptive least
  mean squares estimation of graph signals,'' \emph{IEEE Trans. on Signal and
  Inf. Proc. over Networks}, vol.~2, no.~4, pp. 555--568, Dec. 2016.

\bibitem{chen2015distributed}
S.~Chen, A.~Sandryhaila, and J.~Kovacevic, ``Distributed algorithm for graph
  signal inpainting,'' in \emph{IEEE International Conference on Acoustics,
  Speech and Signal Processing}, Brisbane, March 2015, pp. 3731--3735.

\bibitem{thanou2015distributed}
D.~Thanou and P.~Frossard, ``Distributed signal processing with graph spectral
  dictionaries,'' in \emph{Proc. of Allerton Conference on Communication,
  Control, and Computing}, Monticello, Sept. 2015, pp. 1391--1398.

\bibitem{wang2015distributed}
X.~Wang, M.~Wang, and Y.~Gu, ``A distributed tracking algorithm for
  reconstruction of graph signals,'' \emph{IEEE Journal of Selected Topics in
  Signal Processing}, vol.~9, no.~4, pp. 728--740, 2015.

\bibitem{Cattivelli-Sayed}
F.~S. Cattivelli and A.~H. Sayed, ``Diffusion {LMS} strategies for distributed
  estimation,'' \emph{IEEE Trans. on Signal Processing}, vol.~58, no.~3, pp.
  1035--1048, March 2010.

\bibitem{lopes2008diffusion}
C.~G. Lopes and A.~H. Sayed, ``Diffusion least-mean squares over adaptive
  networks: Formulation and performance analysis,'' \emph{IEEE Transactions on
  Signal Processing}, vol.~56, no.~7, pp. 3122--3136, 2008.

\bibitem{takahashi2010diffusion}
N.~Takahashi, I.~Yamada, and A.~H. Sayed, ``Diffusion least-mean squares with
  adaptive combiners: Formulation and performance analysis,'' \emph{IEEE
  Transactions on Signal Processing}, vol.~58, no.~9, pp. 4795--4810, 2010.

\bibitem{Chen-Sayed}
J.~Chen and A.~H. Sayed, ``Diffusion adaptation strategies for distributed
  optimization and learning over networks,'' \emph{IEEE Trans. on Signal
  Processing}, vol.~60, no.~8, pp. 4289--4305, August 2012.

\bibitem{dilorenzo2013sparse}
P.~Di~Lorenzo and A.~H. Sayed, ``Sparse distributed learning based on diffusion
  adaptation,'' \emph{IEEE Transactions on Signal Processing}, vol.~61, no.~6,
  pp. 1419--1433, 2013.

\bibitem{sayed2014adaptation}
A.~Sayed, ``Adaptation, learning, and optimization over networks,''
  \emph{Foundations and Trends{\textregistered} in Machine Learning}, vol.~7,
  no. 4-5, pp. 311--801, 2014.

\bibitem{chen2014multitask}
J.~Chen, C.~Richard, and A.~H. Sayed, ``Multitask diffusion adaptation over
  networks,'' \emph{IEEE Transactions on Signal Processing}, vol.~62, no.~16,
  pp. 4129--4144, 2014.

\bibitem{chen2015diffusion}
------, ``Diffusion lms over multitask networks,'' \emph{IEEE Transactions on
  Signal Processing}, vol.~63, no.~11, pp. 2733--2748, 2015.

\bibitem{chen2017multitask}
------, ``Multitask diffusion adaptation over networks with common latent
  representations,'' \emph{To appear in IEEE Journal of Selected Topics in
  Signal Processing}, 2017.

\bibitem{Chung1997}
F.~R.~K. Chung, \emph{Spectral Graph Theory}.\hskip 1em plus 0.5em minus
  0.4em\relax American Mathematical Society, 1997.

\bibitem{gadde2014active}
A.~Gadde, A.~Anis, and A.~Ortega, ``Active semi-supervised learning using
  sampling theory for graph signals,'' in \emph{Proceedings of the 20th ACM
  SIGKDD international conference on Knowledge discovery and data mining}, New
  York, Aug. 2014, pp. 492--501.

\bibitem{kempe2004decentralized}
D.~Kempe and F.~McSherry, ``A decentralized algorithm for spectral analysis,''
  in \emph{Proceedings of the ACM symposium on Theory of computing}, Chicago,
  June 2004, pp. 561--568.

\bibitem{bertrand2013seeing}
A.~Bertrand and M.~Moonen, ``Seeing the bigger picture: How nodes can learn
  their place within a complex ad hoc network topology,'' \emph{IEEE Signal
  Processing Magazine}, vol.~30, no.~3, pp. 71--82, 2013.

\bibitem{di2014distributed}
P.~Di~Lorenzo and S.~Barbarossa, ``Distributed estimation and control of
  algebraic connectivity over random graphs,'' \emph{IEEE Transactions on
  Signal Processing}, vol.~62, no.~21, pp. 5615--5628, 2014.

\bibitem{dilorenzo2014diffusion}
P.~Di~Lorenzo, ``Diffusion adaptation strategies for distributed estimation
  over {Gaussian Markov} random fields,'' \emph{IEEE Transactions on Signal
  Processing}, vol.~62, no.~21, pp. 5748--5760, 2014.

\bibitem{fernandez2012novel}
J.~Fern{\'a}ndez-Bes, J.~A. Azpicueta-Ruiz, M.~T. Silva, and
  J.~Arenas-Garc{\'\i}a, ``A novel scheme for diffusion networks with
  least-squares adaptive combiners,'' in \emph{2012 IEEE International Workshop
  on Machine Learning for Signal Processing}, Santander, Sept. 2012, pp. 1--6.

\bibitem{lopes2014towards}
C.~G. Lopes, L.~F. Chamon, and V.~H. Nascimento, ``Towards spatially universal
  adaptive diffusion networks,'' in \emph{IEEE Global Conference on Signal and
  Information Processing}, Atlanta, Dec. 2014, pp. 803--807.

\bibitem{di2013bio}
P.~Di~Lorenzo, S.~Barbarossa, and A.~H. Sayed, ``Bio-inspired decentralized
  radio access based on swarming mechanisms over adaptive networks,''
  \emph{IEEE Transactions on Signal Processing}, vol.~61, no.~12, pp.
  3183--3197, 2013.

\bibitem{di2013distributed}
------, ``Distributed spectrum estimation for small cell networks based on
  sparse diffusion adaptation,'' \emph{IEEE Signal Processing Letters},
  vol.~20, no.~12, pp. 1261--1265, 2013.

\bibitem{chouvardas2011adaptive}
S.~Chouvardas, K.~Slavakis, and S.~Theodoridis, ``Adaptive robust distributed
  learning in diffusion sensor networks,'' \emph{IEEE Transactions on Signal
  Processing}, vol.~59, no.~10, pp. 4692--4707, 2011.

\bibitem{Barb-Sard-Dilo}
S.~Barbarossa, S.~Sardellitti, and P.~{Di Lorenzo}, \emph{Distributed Detection
  and Estimation in Wireless Sensor Networks}.\hskip 1em plus 0.5em minus
  0.4em\relax Academic Press Library in Signal Processing, 2014, vol.~2, pp.
  329--408.

\bibitem{sayed2011adaptive}
A.~H. Sayed, \emph{Adaptive filters}.\hskip 1em plus 0.5em minus 0.4em\relax
  John Wiley \& Sons, 2011.

\bibitem{horn2012matrix}
R.~A. Horn and C.~R. Johnson, \emph{Matrix analysis}.\hskip 1em plus 0.5em
  minus 0.4em\relax Cambridge university press, 2012.

\bibitem{xiao2007distributed}
L.~Xiao, S.~Boyd, and S.-J. Kim, ``Distributed average consensus with
  least-mean-square deviation,'' \emph{Journal of Parallel and Distributed
  Computing}, vol.~67, no.~1, pp. 33--46, 2007.

\bibitem{chepuri2016subsampling}
S.~P. Chepuri and G.~Leus, ``Subsampling for graph power spectrum estimation,''
  in \emph{IEEE Sensor Array and Multichannel Signal Processing Workshop
  (SAM)}, Rio de Janeiro, July 2016.

\bibitem{olfati2004consensus}
R.~Olfati-Saber and R.~M. Murray, ``Consensus problems in networks of agents
  with switching topology and time-delays,'' \emph{IEEE Transactions on
  Automatic Control}, vol.~49, no.~9, pp. 1520--1533, 2004.

\bibitem{chepuri2015sparsity}
S.~P. Chepuri and G.~Leus, ``Sparsity-promoting sensor selection for non-linear
  measurement models,'' \emph{IEEE Transactions on Signal Processing}, vol.~63,
  no.~3, pp. 684--698, 2015.

\bibitem{shamaiah2010greedy}
M.~Shamaiah, S.~Banerjee, and H.~Vikalo, ``Greedy sensor selection: Leveraging
  submodularity,'' in \emph{IEEE conference on decision and control}, Atlanta,
  Georgia, USA, December 2010, pp. 2572--2577.

\bibitem{silvester2000determinants}
J.~R. Silvester, ``Determinants of block matrices,'' \emph{The Mathematical
  Gazette}, vol.~84, no. 501, pp. 460--467, 2000.

\end{thebibliography}

\begin{IEEEbiography}[{\includegraphics[scale=0.035]{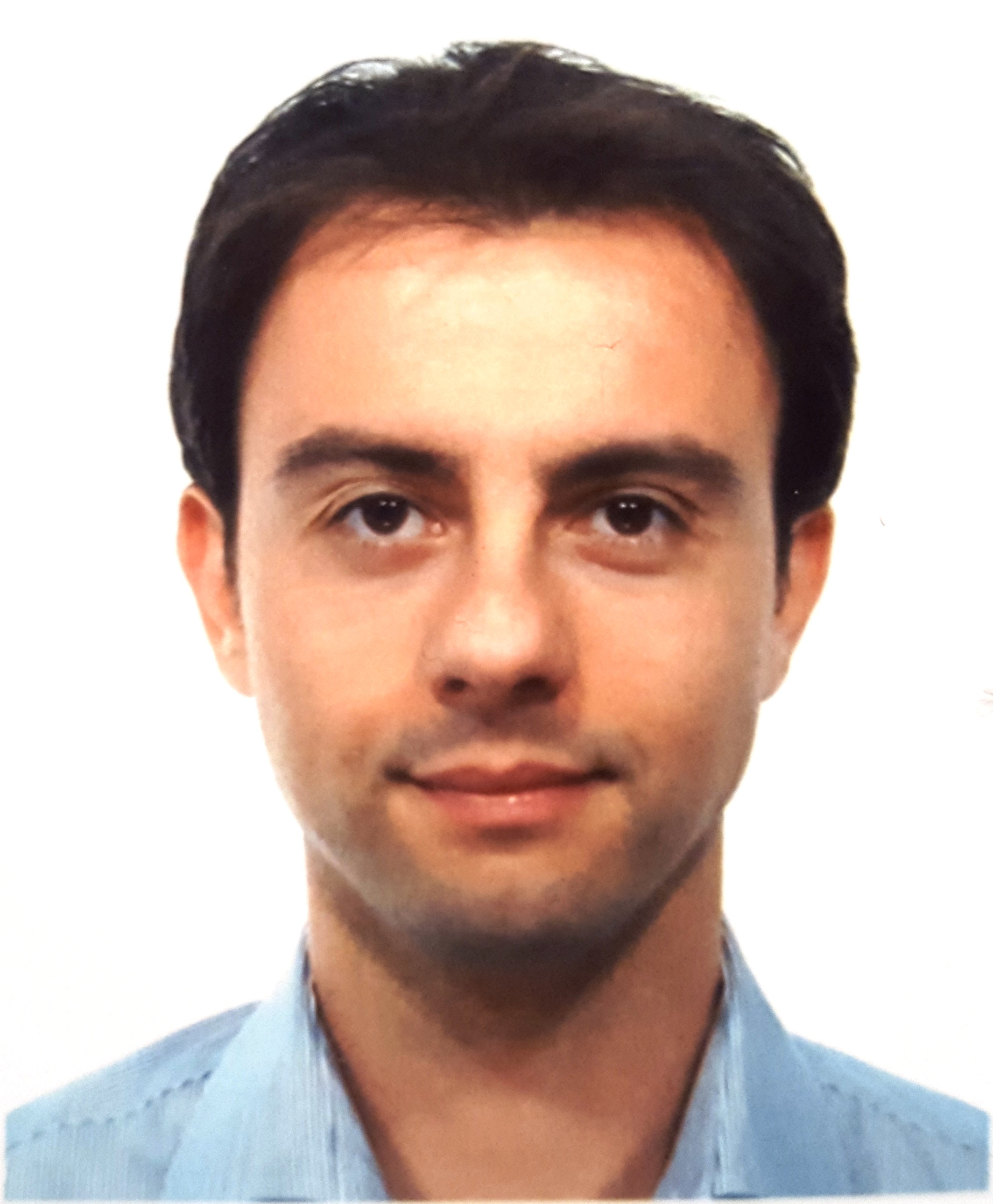}}]{Paolo Di Lorenzo}
(S'10-M'13)  received the M.Sc. degree in 2008 and the Ph.D. in electrical engineering in 2012, both from University of Rome ``Sapienza,'' Italy. He is an assistant Professor in the Department of Engineering, University of Perugia, Italy. During 2010 he held a visiting research appointment in the Department of Electrical Engineering, University of California at Los Angeles (UCLA). He has participated in the European research project FREEDOM, on femtocell networks, SIMTISYS, on moving target detection through satellite constellations, and TROPIC, on distributed computing, storage and radio resource allocation over cooperative femtocells. His current research interests are in signal processing theory and methods, distributed optimization, adaptation and learning over networks, and graph signal processing. He is currently an Associate Editor of the \textit{Eurasip Journal on Advances in Signal Processing}. He received three best student paper awards, respectively at IEEE SPAWC'10, EURASIP EUSIPCO'11, and IEEE CAMSAP'11, for works in the area of signal processing for communications and synthetic aperture radar systems. He is also recipient of the 2012 GTTI (Italian national group on telecommunications and information theory) award for the Best Ph.D. Thesis in information technologies and communications.
\end{IEEEbiography}

\vspace{-.85cm}
\begin{IEEEbiography}[{\includegraphics[scale=0.125]{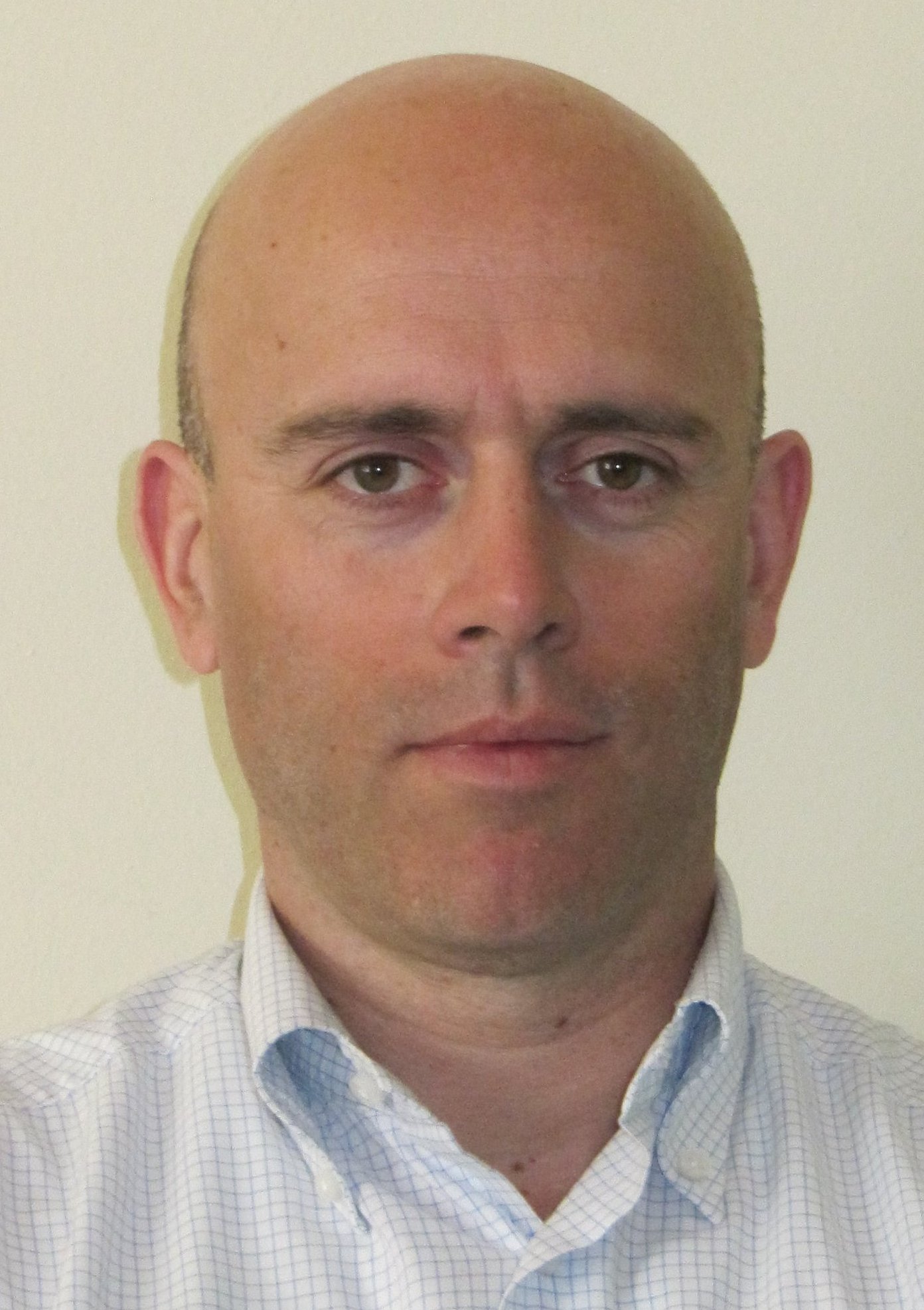}}]{Paolo Banelli}
(S'90-M'99) received the Laurea degree (cum laude) in electronics engineering and the Ph.D. degree in telecommunications from the University
of Perugia, Italy, in 1993 and 1998, respectively. In 2005, he was appointed as Associate Professor in the Department of Electronic and Information Engineering, University of Perugia, where he has been an Assistant Professor since 1998. In 2001, he joined the SpinComm group, as a Visiting Researcher, in the ECE Department, University of Minnesota, Minneapolis. His research interests include signal processing for wireless communications, with emphasis on multicarrier transmissions, signal processing for biomedical applications, spectrum sensing for cognitive radio, waveform design for 5G communications, and recently graph signal processing. He was a Member (2011-2013) of the IEEE Signal Processing Society's Signal Processing for Communications and Networking Technical Committee. In 2009, he was a General Cochair of the IEEE International Symposium on Signal Processing Advances for Wireless Communications. He currently serves as an Associate Editor of the IEEE
TRANSACTIONS ON SIGNAL PROCESSING and the \textit{EURASIP Journal on Advances in Signal Processing}.
\end{IEEEbiography}

%\vspace{-.9cm}
\begin{IEEEbiography}[{\includegraphics[scale=0.12]{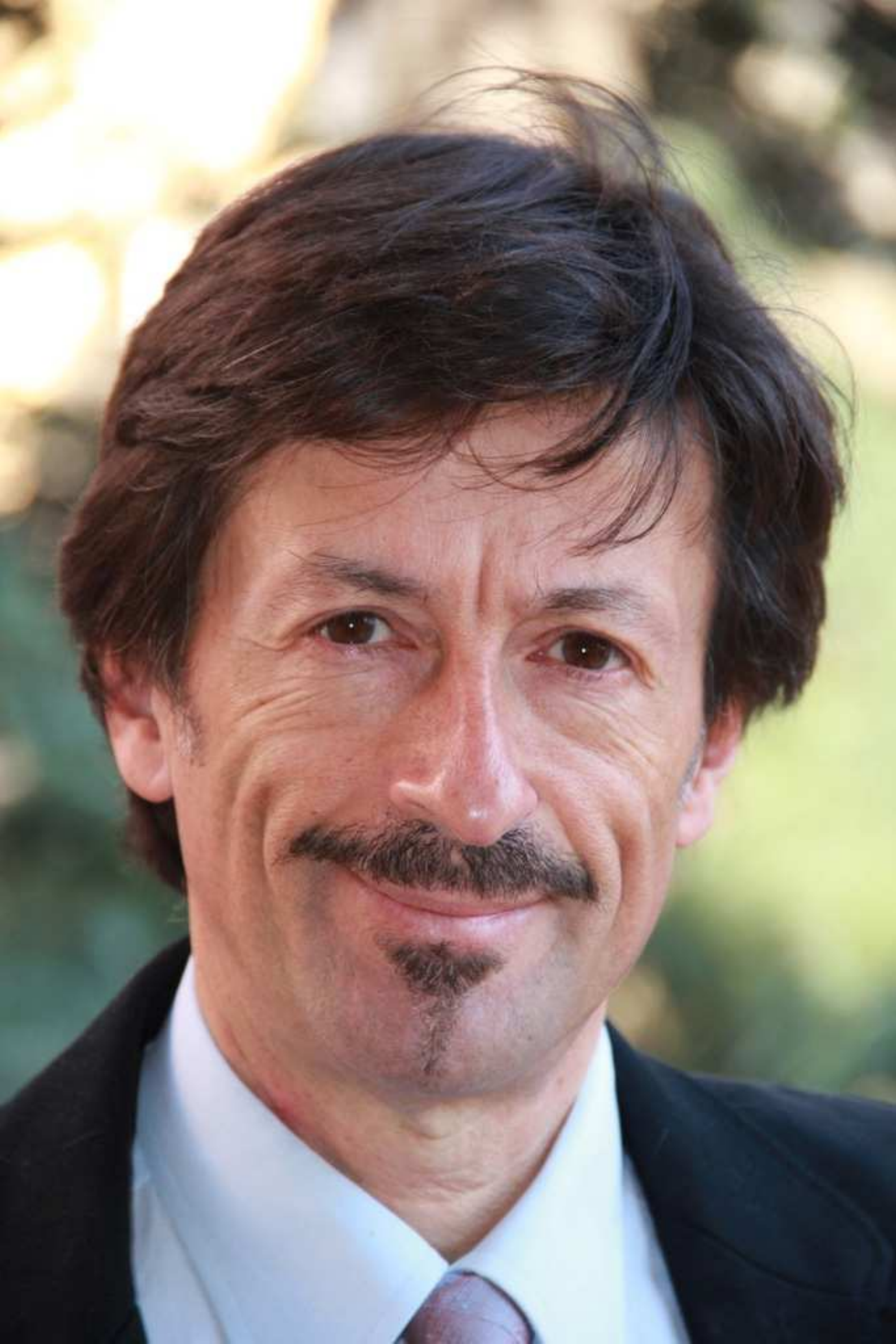}}]{Sergio Barbarossa}
(S'84-M'88-F'12) received the M.Sc. and Ph.D. degrees in electrical engineering from Sapienza University of Rome, Italy, in 1984 and 1988, respectively. He has held positions as a Research Engineer with Selenia SpA (1984-1986) and with the Environmental Institute of Michigan (1988), as a Visiting Professor at the University of Virginia (1995 and 1997) and at the University of Minnesota (1999). Currently, he is a Full Professor with the University of Rome ``Sapienza.'' He received the 2010 EURASIP Technical Achievements Award and the 2000 and 2014 IEEE Best Paper Awards from the IEEE Signal Processing Society. He served as IEEE Distinguished Lecturer in 2012-2013. He is the author of a research monograph titled Multiantenna Wireless Communication Systems. He has been the Scientific Coordinator of various European projects on wireless sensor networks, femtocell networks, and mobile cloud computing. Prof. Barbarossa is an IEEE Fellow and an EURASIP Fellow. He has been the General Chairman of the IEEE Workshop on Signal Processing Advances in Wireless Communications (SPAWC) in 2003, and the Technical Co-Chair of SPAWC in 2013. He served as an Associate Editor for the IEEE TRANSACTIONS ON SIGNAL PROCESSING (1998-2000 and 2004-2006) and the IEEE SIGNAL PROCESSING MAGAZINE. He is currently an Associate Editor of the IEEE TRANSACTIONS ON SIGNAL AND INFORMATION PROCESSING OVER NETWORKS.  He is the coauthor of papers that received the Best Student Paper Award at ICASSP 2006, SPAWC 2010, EUSIPCO 2011, and CAMSAP 2011. His current research interests include topological data analysis, signal processing over graphs, mobile-edge computing, and 5G networks.
\end{IEEEbiography}

%\vspace{-.8cm}
\begin{IEEEbiography}[{\includegraphics[scale=0.07]{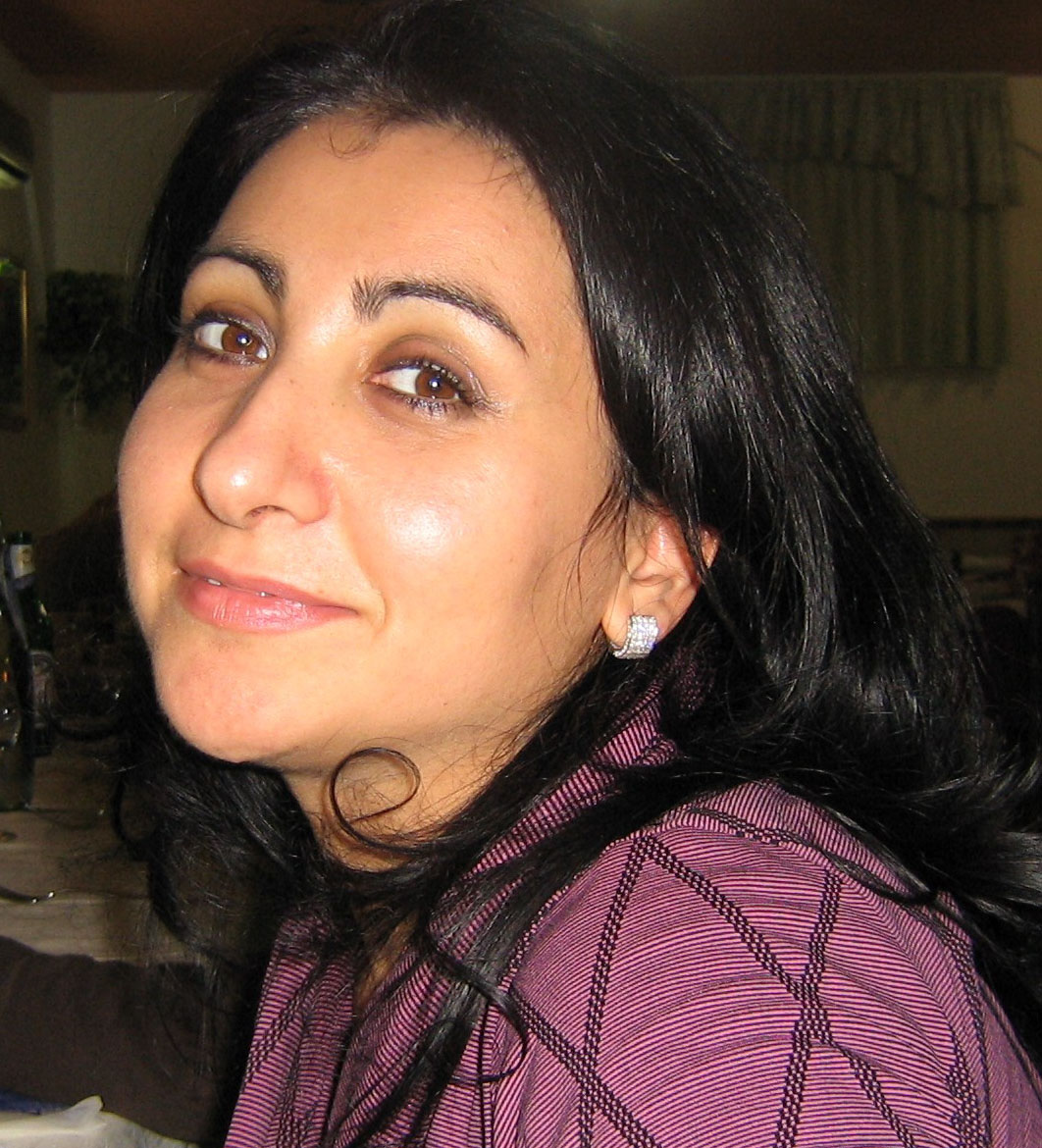}}]{Stefania Sardellitti}
(M'12) received the  M.Sc. degree in Electronic Engineering from the University of Rome ``Sapienza,'' Italy, in 1998 and   the  Ph.D. degree  in Electrical and Information Engineering from the University of Cassino, Italy, in 2005. Since 2005 she is an appointed professor of digital communications at the University of  Cassino, Italy. She is  a research assistant at the  Department of Information, Electronics and Telecommunications, University of Rome,  Sapienza,  Italy. She received the 2014 IEEE Best Paper Award from the IEEE Signal Processing Society. She has participated in the European project WINSOC (on wireless sensor networks) and in the European projects  FREEDOM (on femtocell networks) and TROPIC, on distributed computing, storage and radio resource allocation over cooperative femtocells. Her research interests are in the area of statistical signal processing, mobile cloud computing, femtocell networks and wireless sensor networks, with emphasis on distributed optimization.
\end{IEEEbiography}

\end{document}